\documentclass[DTMColor]{ROB-New}
\usepackage{subcaption}
\begin{document}

\authormark{Cambridge Authors}

\articletype{RESEARCH ARTICLE}

\jnlPage{1}{7}
\jyear{2021}
\jdoi{10.1017/xxxxx}

\title{Entropy-Controlled Intrinsic Motivation Reinforcement Learning for Quadruped Robot Locomotion in Complex Terrains}

\author[1,2]{Wanru Gong\hyperlink{corr}{a}}
\address[1]{Faculty of Information Technology, Beijing University of Technology, No.100 Ping Le Yuan, Chaoyang District, Beijing, 100124, PR China}
\address[2]{Faculty Nuclear Industry X Intelligence Laboratory, Beijing University of Technology, No.100 Ping Le Yuan, Chaoyang District, Beijing, 100124, PR China}

\author[3]{Xinyi Zheng\hyperlink{corr}{a}}
\address[3]{Department of Computer Science, University of Bristol, Beacon House, Queens Rd, BS8 1QU, Bristol, England}

\author[3]{Yuan Hui\hyperlink{corr}{}}

\author[1,2]{Zhongjun Li}
\author[1,2]{Weiqiang Wang}
\author[1,2]{Xiaoqing Zhu\hyperlink{corr}{*}}
\address{\hypertarget{corr}{*}Corresponding author. \email{alex.zhuxq@bjut.edu.cn}}

\address{\hypertarget{corr}{a}These authors contributed equally to this article as the first author}
\received{xx xxx xxx}
\revised{xx xxx xxx}
\accepted{xx xxx xxx}


\abstract{

Learning is the basis of both biological and artificial systems when it comes to mimicking intelligent behaviors. From the classical PPO (Proximal Policy Optimization), there is a series of deep reinforcement learning algorithms which are widely used in training locomotion policies for quadrupedal robots because of their stability and sample efficiency. However, among all these variants, experiments and simulations often converge prematurely, leading to suboptimal locomotion and reduced task performance. Therefore, in this paper, we introduce Entropy-Controlled Intrinsic Motivation (ECIM), an entropy-based reinforcement learning algorithm in contrast with the PPO series, that can reduce premature convergence by combining intrinsic motivation with adaptive exploration.

For experiments, in order to parallel with other baselines, we chose to apply it in Isaac Gym across six terrain categories: upward slopes, downward slopes, uneven rough terrain, ascending stairs, descending stairs, and flat ground as widely used. For comparison, our experiments consistently achieve better performance: task rewards increase by 4–12\%, peak body pitch oscillation is reduced by 23–29\%, joint acceleration decreases by 20–32\%, and joint torque consumption declines by 11–20\%. Overall, our model ECIM, by combining entropy control and intrinsic motivation control, achieves better results in stability across different terrains for quadrupedal locomotion, and at the same time reduces energetic cost and makes it a practical choice for complex robotic control tasks.

}

\maketitle

\section{Introduction}
In recent years, mobile robots have garnered significant attention for their ability to autonomously carry out tasks across diverse environments~\cite{10.1115/1.4039394}. Among these, quadruped robots—equipped with bio-inspired legged architectures~\cite{10.1108/IR-11-2016-0310} or legged walking robots~\cite{CHEN2024105464}—exhibit remarkable agility and environmental adaptability, enabling stable and animal-like locomotion over complex and dynamic terrains~\cite{10552939,ms-9-1-2018}. Furthermore, compared to other legged robot configurations, quadrupeds strike an advantageous balance between structural stability and controllability, rendering them particularly well-suited for a wide range of real-world applications.
Reinforcement learning (RL) enables agents to learn environment interaction through trial and error, aiming to maximize cumulative reward or achieve predefined goals \cite{712192, doi:10.1287/ijoc.1080.0305}. Through this, RL makes agents learn how to learn.
In recent years, reinforcement learning (RL) has driven substantial progress across a wide range of domains, including game theory~\cite{Szita2012}, robotic control~\cite{DBLP:journals/corr/abs-1812-03201,7758092}, autonomous driving~\cite{9351818,DBLP:journals/corr/Shalev-ShwartzS16a}, recommendation systems~\cite{CHEN2023110335}, and natural language processing~\cite{DBLP:journals/corr/HeCHGLDO15,FERREIRA2015256}.
In RL-based robotic control frameworks \cite{DBLP:journals/corr/abs-1812-00568}, agents observe the environment and take actions based on their current policy.

Proximal Policy Optimization (PPO)~\cite{DBLP:journals/corr/SchulmanWDRK17} is a widely used policy gradient method. It stabilizes training by clipping probability ratios, thereby approximately enforcing a trust-region constraint. By constraining the size of each policy update, PPO helps reduce the instability often seen in standard policy gradient methods. It has been widely used in robotics tasks~\cite{DBLP:journals/corr/abs-2109-11978} and more recently in multi-model
agent interaction~\cite{11044689}. In quadrupedal locomotion, however, several studies have shown that PPO
still faces clear limitations, particularly in achieving stable gait transitions and consistent tracking performance~\cite{9765709,4054408}. The learned policy often converges to suboptimal gaits, getting stuck in poor local optima~\cite{sutton2018reinforcement}. 
This issue is particularly pronounced in complex, uneven terrain, where learned policies often converge to suboptimal gaits and become trapped in poor local optima~\cite{sutton2018reinforcement}. For  this, we improve a framework based on entropy.

Incorporating an entropy bonus encourages exploration and helps prevent premature convergence~\cite{DBLP:journals/corr/abs-1801-01290}.
Too much entropy induces excessive randomness in actions, which harms control accuracy and compromises system stability~\cite{ziebart2008maximum}. Conversely, too little entropy reduces behavioral diversity, leading to premature convergence and degraded policy performance. To balance exploration and exploitation, we propose dynamically adjusting the entropy coefficient during training. Through this our policy can adjust by itself.
Motion smoothness also plays a critical role in quadrupedal locomotion, as legged systems are highly sensitive to abrupt changes in joint trajectories~\cite{10354849,Wang_Wang_Zhang_Dai_2022}. Discontinuous actions can destabilize the robot’s center of mass, leading to imbalance or falls~\cite{6696874,6518581}. We mitigate balance loss from discontinuous commands by constraining action changes across time steps, ensuring physically plausible motion. Smooth transitions between gaits allow the robot to adapt rapidly and stably to terrain changes or pushes. for this, robots can run smoothly in different environments. In many reinforcement learning problems, extrinsic rewards are sparse or delayed~\cite{DBLP:journals/corr/abs-1810-12894,DBLP:journals/corr/BellemareSOSSM16}. To address this, intrinsic rewards are commonly used to promote exploration. 

Existing methods typically optimize exploration, motion smoothness, and curiosity in isolation, ignoring their dependencies. 
Although PPO enables strong locomotion performance~\cite{pmlr-v164-rudin22a}, policies frequently suffer from gait instability and excessive body sway in unstructured settings. In environments which are not flat, policy often become worse.
We propose a framework that integrates curiosity-driven exploration, motion smoothness, and adaptive entropy to overcome the limitations of decoupled designs.
To bridge this gap, we propose ECIM, a unified extension of PPO that jointly optimizes adaptive entropy scheduling, explicit action smoothness regularization, and terrain-aware intrinsic rewards.

Our approach supports autonomous gait modulation and dynamic balance maintenance in diverse, unstructured environments. Robot can get great policy through our approach. The Entropy-Controlled Intrinsic Motivation (ECIM) algorithm forms the foundation of our framework. Built on PPO, it integrates three complementary mechanisms:
\begin{itemize}
    \item \textbf{Adaptive Entropy-Controlled Policy Optimization Mechanism (AECPOM):} It adapts the entropy coefficient in response to empirical returns, promoting broad exploration initially and increasingly favoring exploitation as the policy stabilizes—striking a balance between discovery and convergence.
    \item \textbf{Motion Continuity Regularization Framework (MCRF):} It imposes joint smoothness constraints across time (to limit abrupt action transitions) and state space (to reduce sensitivity to minor perceptual perturbations), yielding more stable gaits and consistent locomotion behavior.
    \item \textbf{Intrinsic Motivation-Driven Exploration Enhancement Mechanism (IMDEEM):} It guides exploration toward states that are both novel and task-relevant—such as traversable regions or stable footholds—thereby improving sample efficiency without sacrificing goal-directednes
\end{itemize}

These components are jointly optimized within a single objective, ensuring that exploration, smoothness, and task relevance mutually reinforce one another.
It boosts learning efficiency, movement smoothness, and exploration quality. Building on the experimental setup from Nikita Rudin et al.~\cite{pmlr-v164-rudin22a}, we  ran a series of controlled simulations in Isaac Gym across six different terrain types. We use the same settings to fairly compare ECIM against PPO baseline. Based on the comparison of experimental results, we can conclude that: ECIM is better than PPO. It has higher rewards, shows less body pitching and has smoother joint movements. The results clearly show that ECIM delivers better control and greater energy efficiency in all kinds of ter rains.

\section{Related Work}
\subsection{Reinforcement Learning}
The goal of RL is to learn a policy to maximize the expected cumulative reward in a Markov decision process (MDP)  framework~\cite{puterman2014markov}.
MDP is defined by the tuple \((S, A, P, R)\), which consists of a set of states \(s \in S\), a set of actions \(a \in A\), transition dynamics \(P(s' \mid s, a)\), and a reward function \(r = R(s, a)\).
In online RL, the agent interacts with the environment to collect trajectory data and improve policy. A trajectory is denoted as \((s_t, a_t, r_t)_{t=0}^{T}\), where \(s_t, a_t,\), and \(t\) represent the state, action, and reward at time step t, respectively, and \(T\) denotes the episode length.

\subsection{The Exploration-exploitation Dilemma}

Balancing exploration and exploitation is a challenge in RL~\cite{thrun1992efficient}. In uncertain environments, agents interactions with the environment and gather information. The dilemma lies in whether the agent should select actions it believes to be optimal based on existing experience or choose actions with high uncertainty. Exploration refers to the agent trying novel actions to discover new states and rewards, while exploitation involves selecting actions known to yield high rewards~\cite{kober2013reinforcement}.

If an agent focuses solely on exploitation, it risks converging to suboptimal policies due to incomplete information, potentially missing better strategies. Conversely, if the agent performs uniform random exploration, it may repeatedly sample low-reward state-action pairs, increasing the number of low-quality samples and wasting time and resources without effectively accumulating rewards. Therefore, reinforcement learning algorithms must balance exploration and exploitation to efficiently explore the environment and reduce uncertainty about it.
\subsection{Detailed Explanation of the PPO Objective}

Our model here based on the traditional and efficient model PPO framework. At the same time, we improves the stability and sample efficient by incorporating a clipping mechanism into the standard policy gradient objective~\cite{sutton2000policy}. 

At each timestep \(t\), the model is given the environment state \(s_t\), the policy network with parameters \(\theta\) outputs a distribution results over actions \(\pi_{\theta}(\cdot\mid s_t)\), from which the agent samples \(a_t\sim\pi_{\theta}(\cdot\mid s_t)\).

\paragraph{Importance Sampling Ratio}
Define the importance sampling ratio as following:
\begin{align}
r_t(\theta)
\;=\;
\frac{\pi_{\theta}(a_t\mid s_t)}
     {\pi_{\theta_{\mathrm{old}}}(a_t\mid s_t)}\,.
\end{align}

This ratio measures how the probability assigned to \((s_t,a_t)\) changes under the new policy compared to the old one.  If \(r_t(\theta)\) is differentiated too far from \(1\), it shows an overly large policy update, which can destabilize training.

\paragraph{Generalized Advantage Estimation}
We employ the Generalized Advantage Estimate (GAE) \(\hat A_t\) to quantify the relative desirability of action \(a_t\) compared to the baseline value function~\cite{schulman2015high}.  A higher \(\hat A_t\) indicates that \(a_t\) yielded a return above the expected value, so the policy should be reinforced in that direction.

\paragraph{Clipped Surrogate Objective}
Rather than use the unconstrained objective \([r_t(\theta)\,\hat A_t]\), PPO introduces clipping to limit policy updates:
\begin{align}
L^{\mathrm{CLP}}(\theta)
\;=\;\min\Bigl(r_t(\theta)\,\hat A_t,\;
\mathrm{clip}\bigl(r_t(\theta),1-\epsilon,1+\epsilon\bigr)\,\hat A_t\Bigr).
\end{align}

If \(r_t(\theta)\) stays within \([\,1-\epsilon,1+\epsilon\,]\), the true ratio is used; once it exits this interval, it is clipped to the boundary. This preserves the gradient direction for small updates and prevents excessively large policy shifts.

\paragraph{Value Function Loss}
PPO jointly learns a state-value function \(V_\theta(s)\) by minimizing the mean-squared error:
\begin{align}
L^{\mathrm{VF}}(\theta)
\;=\;
\Bigl[\bigl(V_\theta(s_t)-V_{\mathrm{target}}(s_t)\bigr)^2\Bigr].
\end{align}

\paragraph{Full PPO Objective}
Putting these components together, the PPO loss is
\begin{align}
L^{\mathrm{PPO}}_t(\theta)
=
L^{\mathrm{CLP}}(\theta)
\;+\;
c_1\,L^{\mathrm{VF}}(\theta).
\end{align}

In our quadruped gait learning experiments, the standard PPO-trained policy frequently converges to suboptimal locomotion patterns, especially in complex environments. We propose ECIM to overcome these.
\section{Methods}
\begin{figure*}[htbp]  
\centering
\includegraphics[width=\linewidth]{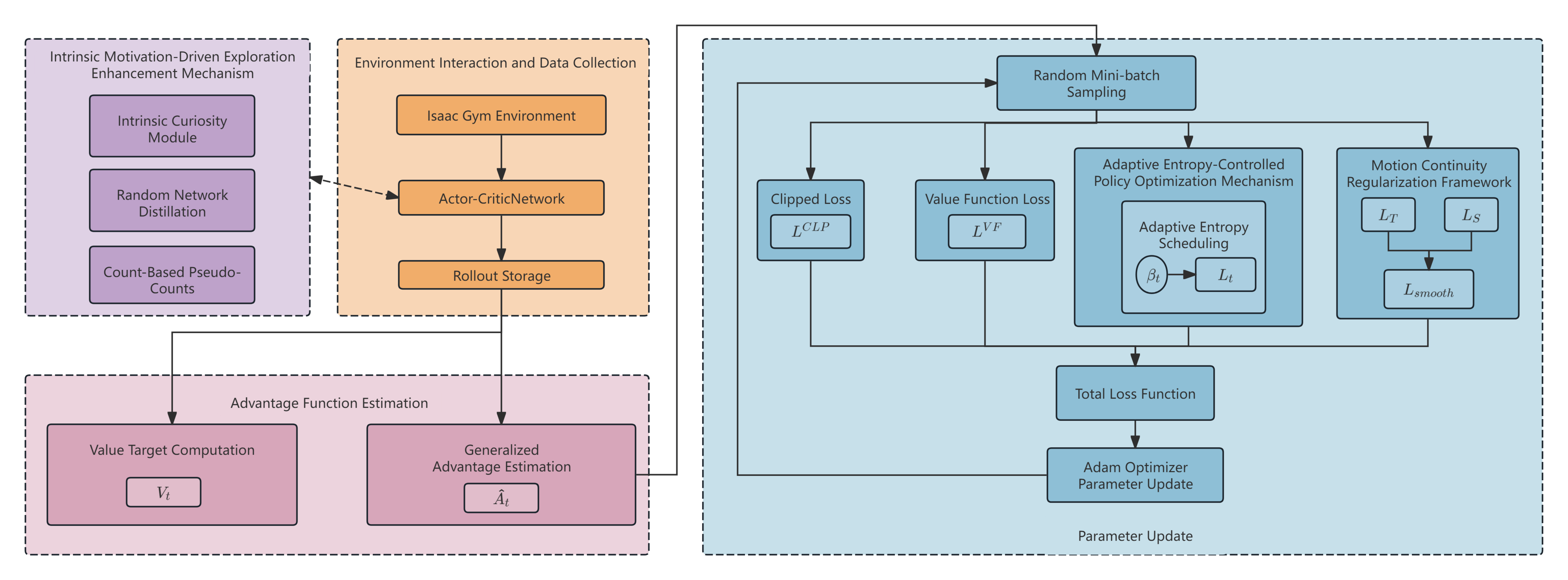}
\caption{Schematic illustration of the ECIM framework, highlighting its three core components.}
\label{fig:Algorithm Flowchart}
\end{figure*}
To solve the challenges of efficient exploration and smooth motion control, We propose a RL framework called Entropy-Controlled Intrinsic Motivation (ECIM). ECIM strike a balance between exploring policy learning and generating stable actions.

We design ECIM (Figure~\ref{fig:Algorithm Flowchart}) around three complementary components: adaptive entropy scheduling, action smoothness regularization, and intrinsic reward guidance to balance exploration, stability, and adaptability. First, we design the entropy term to decay not on policy performance. Second, we impose smoothness regularization on the action sequences. Finally, we us the intrinsic reward guidance component augments sparse extrinsic feedback with curiosity. 
By jointly improving exploration efficiency and action smoothness, ECIM enables agile and stable navigation in complex environments. In this section, we present the complete ECIM architecture.

\subsection{Adaptive Entropy Regularization: Achieving a Balance between Exploration and Exploitation}

During early training, To enhance exploration, we encourage policies to maintain high randomness. We allow the robot to explore various actions. As  learning progresses, the policy gradually transitions from random exploration to more deterministic utilization. which guides the policy toward ultimate convergence. We introduce a time-dependent weighting factor \(\beta_t\). We apply large values of \(\beta_t\) to amplify entropy-based incentives for diverse behavior at first. \(\beta_t\) will gradual decay reduces exploratory pressure as the policy matures.

\begin{align}
R_t      &= \frac{1}{\tau}\sum_{k=0}^{\tau-1} G_{t-k}, \\
\beta_t &= \beta_{\max}\,\frac{R_t}{R_{\max}}. 
\end{align}

We define \(G_{t}\) as the discounted cumulative return, use \(\tau\) to specify the moving-average window, and set \(R_{\max}\) based on the maximum possible episode return under optimal control. The per-step
objective thus becomes:
\begin{equation}\label{eq:adaptive-entropy}
L_t(\theta) \;=\; L^{\mathrm{PPO}}_t(\theta)
\;-\;
\beta_t\,\mathbb{E}\bigl[\mathcal{H}(\pi_{\theta}(\cdot\mid s_t))\bigr].
\end{equation}

We design the dynamic scheduling to initiate policy exploration in the early stages, then enhance the policy.

\begin{figure*}[htbp]  
\centering
\includegraphics[width=0.6\linewidth]{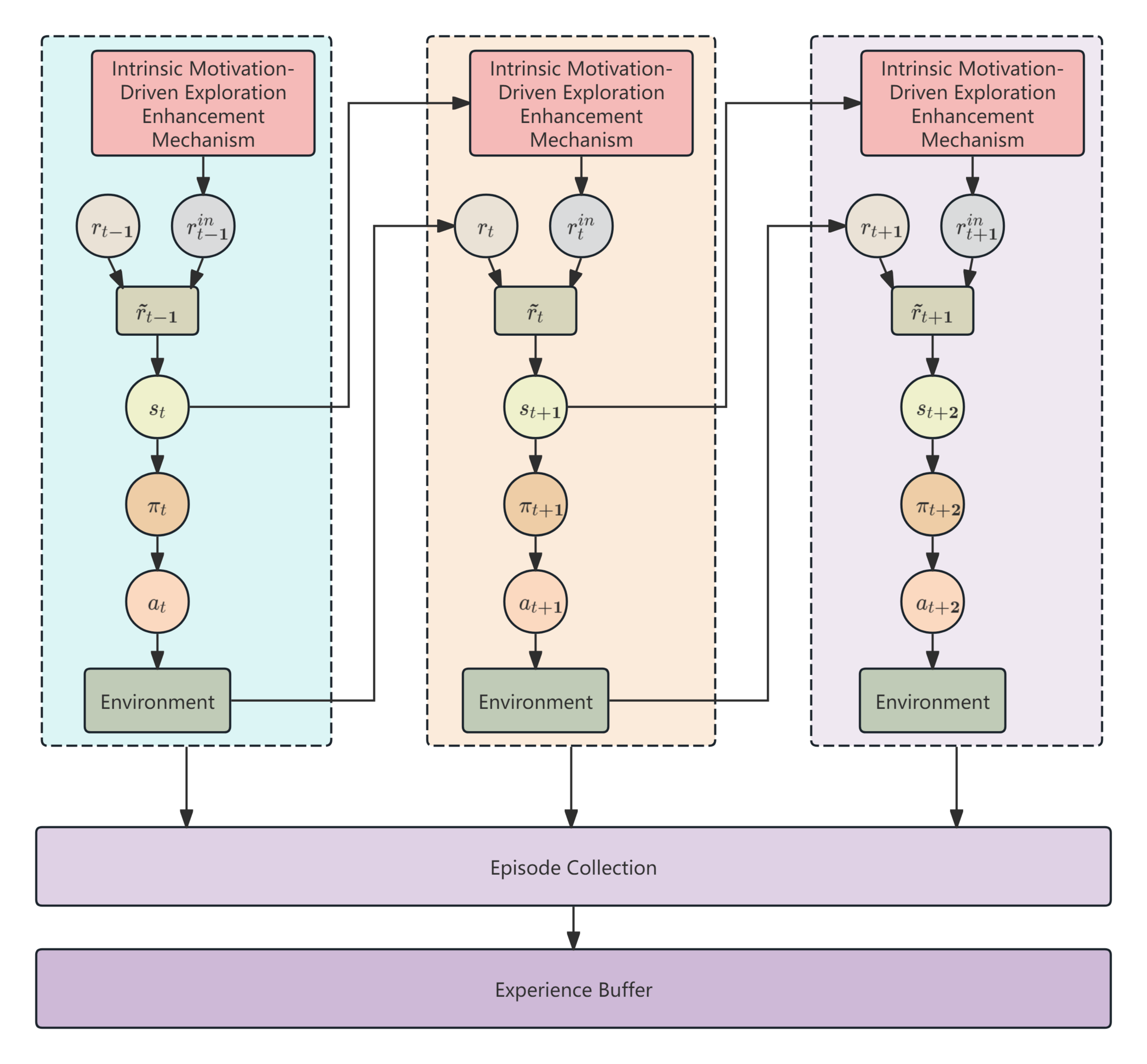}
\caption{Data update flowchart.}
\label{fig:Data Update}    
\end{figure*}

\subsection{Motion Continuity Regularization for Quadrupeds: Ensuring Smoothness and Stability}

The motion of a robot is continuous. To ensure system stability its dynamic behavior must satisfy constraints~\cite{6696874}. In our experiments, we observed that action sequences containing sharp transitions or high-frequency components frequently led to loss of balance and unstable locomotion.

To address these challenges, we introduced two complementary regularization functions to enhance the stability of the learning strategy:
\begin{align}
L_T &= \mathbb{E}_{s_t\sim\mathcal{D}}\bigl[\|\pi_{\theta}(s_{t+1}) - \pi_{\theta}(s_t)\|^2\bigr],\\
L_S &= \mathbb{E}_{s\sim\mathcal{D},\,s'\sim\mathcal{N}(s,\sigma)}\bigl[\|\pi_{\theta}(s') - \pi_{\theta}(s)\|^2\bigr].
\end{align}

\(L_T\) prevents the robot from sudden change. We design \(L_S\) to enhance robustness against input noise. We incorporate these as:
\begin{align}\label{eq:ppo-loss}
L_{\mathrm{smooth}} \;=\; \lambda_T\,L_T \;+\; \lambda_S\,L_S,
\end{align}
$\lambda_T$ and $\lambda_S$ balance time smoothness and space smoothness. 
Robot generate smoother control trajectories and improve system stability by incorporating \(L_{\mathrm{smooth}}\).
\begin{figure*}[htbp] 
\centering
\includegraphics[width=\linewidth]{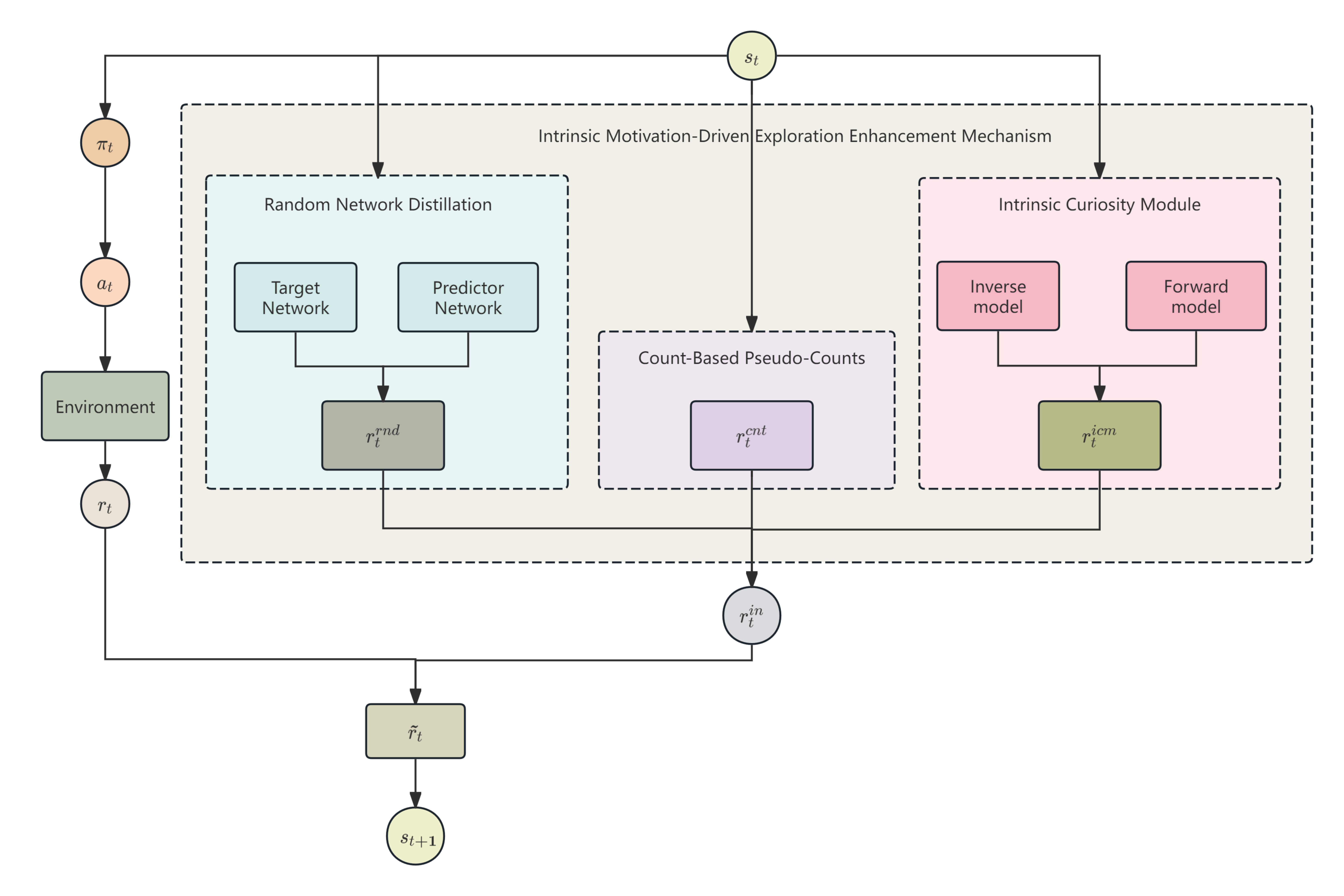}
\caption{Intrinsic Motivation-Driven Exploration Enhancement Mechanism.}
\label{fig:Intrinsic}   
\end{figure*}

\subsection{Intrinsic Rewards for Improved Exploration–Exploitation and Policy Diversity}

The data update process of the ECIM algorithm is showen in Figure~\ref{fig:Data Update}. We introduces an intrinsic reward-guided mechanism to address the inherent exploration-exploitation dilemma in reinforcement learning. It motivates the robot to explore a broader state space. It directs the robot to visit under-explored or information-rich areas in the environment. The intrinsic reward is dynamically adjusted. By prompting the agent to explore in more varied ways, our approach not only enriches policy diversity but also keeps learning stable and convergent. As shown in Figure~\ref{fig:Intrinsic}, the intrinsic
reward-guided mechanism consists of three components:

\subsubsection{Intrinsic Curiosity Module}

Standard extrinsic rewards may be sparse or delayed, which hinders efficient exploration. We equip an intrinsic motivation signal based on prediction error to address this.
Specifically, we learn:
\begin{itemize}
  \item \textbf{Inverse model} \(g_{\phi}\!\bigl(\phi(s_t),\phi(s_{t+1})\bigr)\to a_t\), which
  encourages the encoder \(\phi\) to capture controllable aspects of the environment.
  \item \textbf{Forward model} \(f_{\psi}\!\bigl(\phi(s_t),a_t\bigr)\to \phi(s_{t+1})\), whose
  prediction error reflects the novelty of the transition.
\end{itemize}

The intrinsic reward is defined as:
\begin{align}
r_t^{\mathrm{icm}}
\;=\;
\eta_{\mathrm{icm}}\,\bigl\|\phi(s_{t+1}) - f_{\psi}(\phi(s_t),a_t)\bigr\|^2.
\end{align}
By jointly training \(\phi\), \(f_{\psi}\), and \(g_{\phi}\), the agent is driven to seek out
state transitions that are difficult to predict, thereby improving coverage of the state
space.

\subsubsection{Random Network Distillation}

As an alternative source of intrinsic motivation that does not require explicit inverse
modeling, we adopt Random Network Distillation. A fixed randomly initialized
“target” network \(f_{\mathrm{tgt}}\) provides embeddings of states, and a trainable
“predictor” network \(f_{\mathrm{pred}}\) attempts to match these embeddings:
\begin{align}
r_t^{\mathrm{rnd}}
\;=\;
\eta_{\mathrm{rnd}}\,\bigl\|f_{\mathrm{tgt}}(s_t) - f_{\mathrm{pred}}(s_t)\bigr\|^2.
\end{align}
Since \(f_{\mathrm{pred}}\) gradually learns to mimic \(f_{\mathrm{tgt}}\), the prediction error
naturally decays for familiar states and remains high for novel ones, guiding
exploration without explicit density modeling.

\subsubsection{Count‐Based Pseudo‐Counts}

To further encourage rare-state visitation in high-dimensional settings, we estimate a
density model \(\rho(s)\) and derive pseudo-counts \(\hat N(s)\):
\begin{align}
\hat N(s)
\;=\;
\frac{\rho(s)\bigl(1-\rho'(s)\bigr)}{\rho'(s)\bigl(1-\rho(s)\bigr)},
\end{align}
where \(\rho'\) is the density after updating on \(s\). The corresponding bonus reward
is:
\begin{align}
r_t^{\mathrm{cnt}}
\;=\;
\frac{\eta_{\mathrm{cnt}}}{\sqrt{\hat N(s_t)}}.
\end{align}
This mechanism systematically boosts the reward for under‐visited regions,
complementing the above curiosity signals.

\subsection{Reinforcement Learning Integrated Optimization Architecture}

Finally, we merge all extrinsic and intrinsic signals into a unified per‐step return:
\begin{align}
\tilde r_t \;=\; r_t 
+ r_t^{\mathrm{icm}} 
+ r_t^{\mathrm{rnd}} 
+ r_t^{\mathrm{cnt}}.
\end{align}
and optimize the overall loss:
\begin{align}
\mathcal{L}(\theta)
=
\mathbb{E}\bigl[L^{\mathrm{PPO}}_t(\theta;\tilde r_t)\bigr]
+ \lambda_T L_T + \lambda_S L_S
- \beta_t\,\mathbb{E}\bigl[\mathcal{H}(\pi_{\theta}(\cdot\mid s_t))].
\end{align}

All hyperparameters \(\{\beta_{\max},\tau,\eta_{\cdot},\lambda_{\cdot}\}\) were optimized via grid search to achieve an optimal balance between efficient exploration, smooth control, and stable convergence. The experimental results section below provides a systematic evaluation of the model variants.


\section{Results}

\begin{figure}[htbp]
\centering
\subfloat[Flat Terrain]{%
  \resizebox*{4cm}{!}{\includegraphics{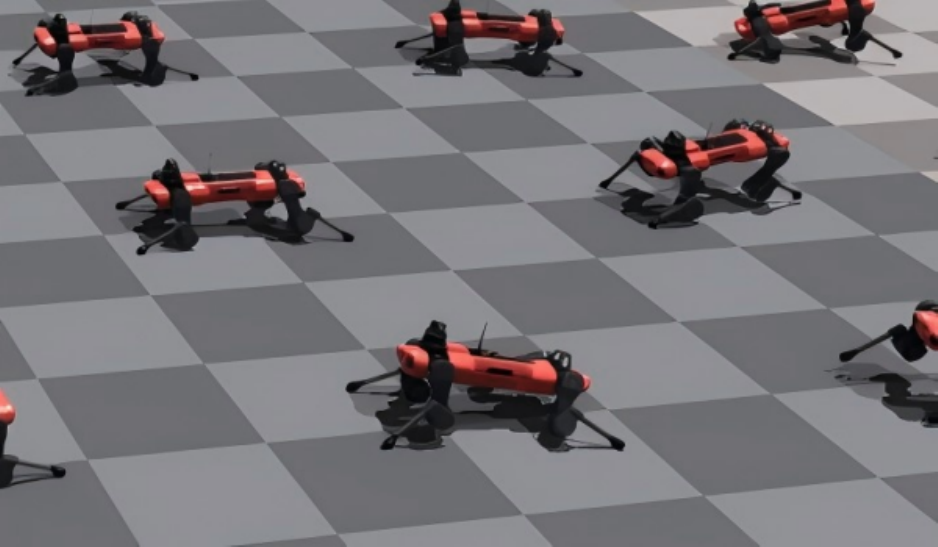}}}\hspace{5pt}
\subfloat[Sloped Terrain]{%
  \resizebox*{4cm}{!}{\includegraphics{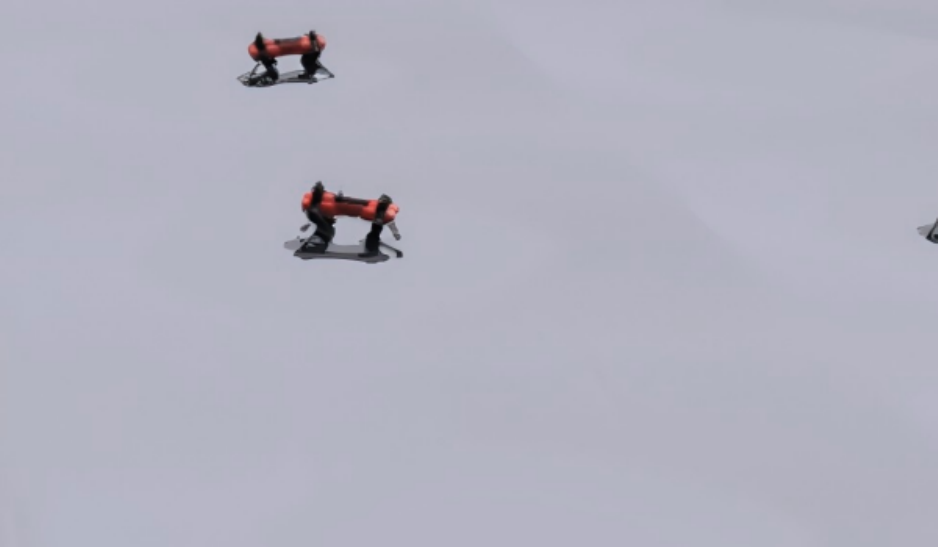}}}\hspace{5pt}
\subfloat[Rough Terrain]{%
  \resizebox*{4cm}{!}{\includegraphics{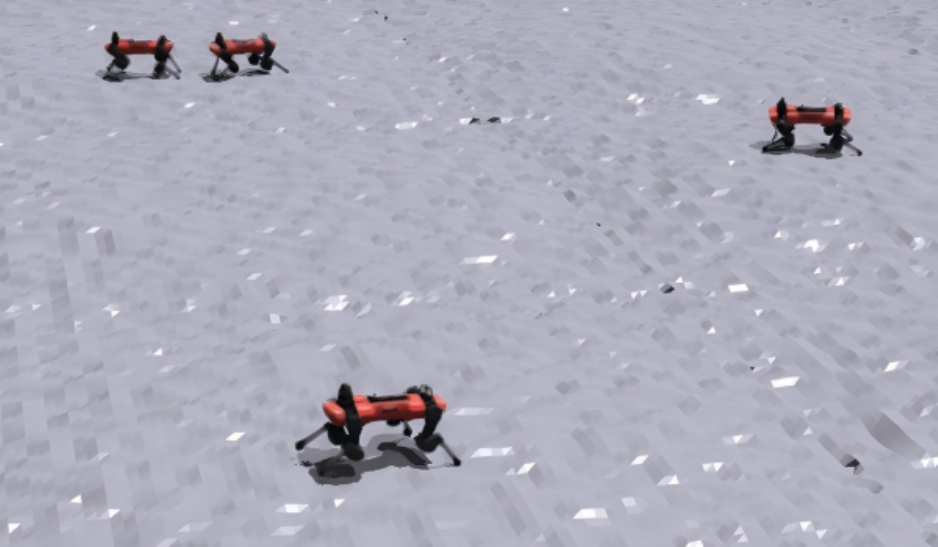}}}\hspace{5pt}
  
\subfloat[Stair-Climbing Terrain]{%
  \resizebox*{4cm}{!}{\includegraphics{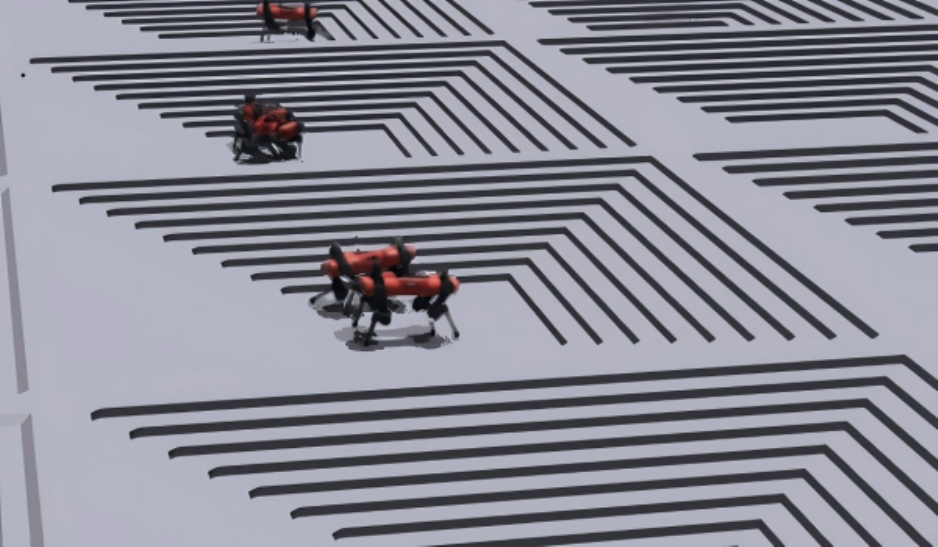}}}\hspace{5pt}
\subfloat[Stair-Descending Terrain]{%
  \resizebox*{4cm}{!}{\includegraphics{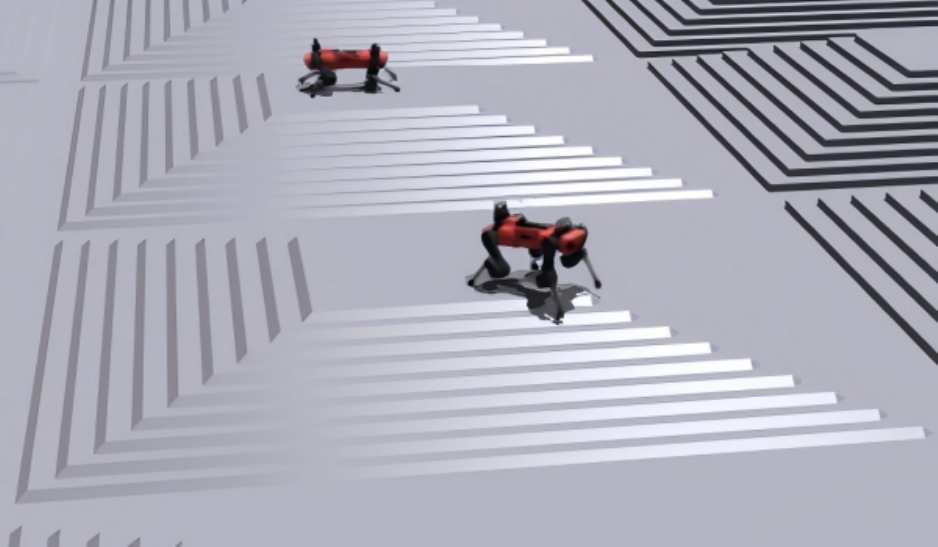}}}\hspace{5pt}
\subfloat[Stepping Stones Terrain]{%
  \resizebox*{4cm}{!}{\includegraphics{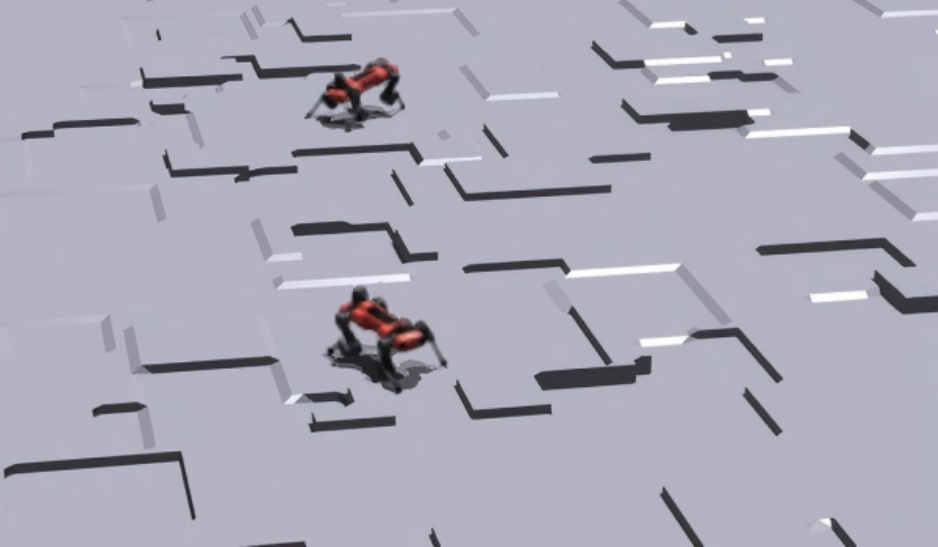}}}
\caption{The robot undergoes training across six different terrain environments: flat, sloped, rough, stair-climbing, stair-descending, and stepping-stones.
The six terrain environments are categorized into three groups based on their complexity: (a) flat terrain (simple); (b)--(c) sloped and rough terrain (medium); and (d)--(f) stair-climbing, stair-descending, and stepping-stones terrain (complex).}
\label{fig:environments}
\end{figure}

\subsection{Experimental Setup}
For our experiment here, we chose the platform Isaac Gym~\cite{DBLP:journals/corr/abs-2108-10470}, which is an advanced physics simulation framework developed by NVIDIA Corporation. It allows large-scale simulation of parallel agents, which provides unprecedented computational capabilities for advancing research in robotics and reinforcement learning. Then, we introduce the ANYmal C as the benchmark platform~\cite{pmlr-v164-rudin22a}. This benchmark platform is a 12-DOF quadruped with three actuated joints per leg, which is widely adopted in prior locomotion work. As shown in Figure~\ref{fig:environments}, we designed six terrains to probe robust locomotion under diverse conditions that approximate real-world scenarios.

In the six terrains of our experiments, these six terrains are chosen for their diversity. The terrains span a range from relatively regular surfaces to highly irregular, rough, or discontinuous structures, covering challenges like uneven ground, height changes and small-area disturbances. All these changes help our model test its ability in adaptive entropy control, motion regularization and intrinsic motivation output stabilization. In combination, these environments provide a systematic testing range for assessing both robustness and adaptability of learned quadruped policies.

Our experiments include 2048 parallel robot instances in total for 2000 iterations, and the performance is evaluated every 50 iterations. The reason we include this large-scale parallelism is to increase the rate of policy sample collection and thus accelerate convergence. Also, our model introduces additional learning signals, which were illustrated previously in the method section, namely entropy control, continuity regularization and intrinsic rewards, and all of these benefit from high-throughput sampling to stabilise the whole interplay. In order to keep its robustness and generalization, this parallel training regime improves sample efficiency and helps ECIM generalize its learned behaviours across the six terrain types while mitigating overfitting.

We adapt the original PPO configuration of Rudin et al. for ANYmal in Isaac Gym as our baseline. For comparison, we use the benchmark source PPO with the same architecture, observation and action spaces. For controlled reasons, we compare the series of benchmarks based on the same pipeline, and our model differs only by the addition of adaptive entropy control, motion continuity regularization and intrinsic motivation components. Under this design, any systematic difference in performance between PPO and ECIM can be attributed to the proposed pipeline rather than differences in architecture, training budget or simulator settings.

\begin{table}[htbp]
\centering
\caption{\label{tab:hyper-parameters}Hyper-parameters used for the training of the tested policy.}
\begin{tabular}{l l }
\toprule

Hyper-parameter & Value \\

\midrule
Batch size & 98304 \\
Mini-bach size & 24576\\
Number of epochs & 5\\
Clip range & 0.2\\
Discount factor & 0.01\\
GAE discount factor & 0.99\\
Desired KL-divergence  & 0.01\\
\hline
\end{tabular}
\end{table}

Table~\ref{tab:hyper-parameters} summarizes the key PPO hyperparameters used for both PPO and ECIM, including batch size, mini-batch size, number of epochs, clipping range, discount factors, and the target KL-divergence. This table follows the widely used configuration from Rudin \emph{et al.}~\cite{pmlr-v164-rudin22a} to ensure the comparability with prior work. At the beginning of each episode, all robot instances are initialized at the geometric center of the corresponding terrain, with identical initial poses, to avoid positional bias and to ensure a fair comparison between methods. Overall, this setup provides a standardized and reproducible experimental environment in which the incremental benefits of the ECIM pipeline over PPO can be rigorously assessed.

\subsection{Evaluation Metrics}

In order to evaluate ECIM in a systematic way, we use a set of metrics that follow the three main design goals of our framework. First, adaptive entropy control (AECPOM) aims to speed up learning and improve task performance. Second, motion continuity regularization (MCRF) targets stable and smooth locomotion. Third, intrinsic motivation for efficient exploration (IMDEEM) focuses on reducing energy usage while still exploring the environment effectively. Besides the standard episode reward, we include several additional metrics that describe learning speed, body stability, motion smoothness and energy-related actuation cost. These metrics are later used both for comparing ECIM with PPO and for understanding how each ECIM module contributes to the final behaviour.

\textbf{Reward}: Reward is the main task-performance metric and measures the total return obtained by the agent during an episode. It is computed from the environment-defined reward function, which typically combines terms for forward velocity, energy consumption, posture stability and other factors. Normally, a higher reward indicates better overall performance in task completion, motion efficiency and stability. The total reward is calculated by:

\begin{align}
R = \sum_{t=1}^{T} r_t,
\end{align}
where $ r_t $ denotes the instantaneous reward at time step $ t $, and $ T $ is the episode length. In our experiments, reward is used as the primary measure to compare PPO and ECIM, and also as one of the metrics for Attribution Gain.

\textbf{Angular velocity loss (\({L}_{\text{ang-xy}}\))}: The pitch angle serves as a fundamental attitude parameter that characterizes the longitudinal stability of a robotic system. Variations in pitch angle reflect the dynamic tilting behavior of the robot’s body in the sagittal plane and directly influence motion smoothness, balance control, load-bearing capability and adaptability to complex terrains. We describe this behavior using the angular velocity vector of the robot base at time step $t$,
\begin{equation}
\boldsymbol{\omega}_t =
\begin{bmatrix}
\omega_{x,t} \\
\omega_{y,t} \\
\omega_{z,t}
\end{bmatrix},
\end{equation}
where $\omega_{x,t}$ and $\omega_{y,t}$ represent the roll and pitch angular velocity components, respectively. The angular velocity energy in the $x$--$y$ plane is computed as
\begin{equation}
\mathcal{L}_{\text{ang-xy}} =
\sum_{t=1}^{N}
\left(
\omega_{x,t}^2 +
\omega_{y,t}^2
\right).
\label{eq:ang_vel_loss}
\end{equation}
where $N$ is the number of time steps. Excessive oscillations in roll and pitch can induce undesired body tilting, shift center of the total mass and increase the likelihood of instability or falling. A lower $\mathcal{L}_{\text{ang-xy}}$ therefore indicates a more stable base motion, which is one of the main targets of MCRF.

\textbf{Joint acceleration}: Joint acceleration is the change of joint velocity over time. It is used to assess how smooth and continuous the robot's movements are. In general, lower joint acceleration means smoother trajectories without sudden start or stop, and often better energy usage. At each simulation step, we compute the joint acceleration as the difference between the current and previous joint velocities divided by the time interval $\Delta t$:
\begin{equation}
\boldsymbol{a}_{t} = 
\frac{\boldsymbol{\dot{q}}_{t} - \boldsymbol{\dot{q}}_{t-1}}{\Delta t},
\label{eq:joint_acc_def}
\end{equation}
where $\boldsymbol{\dot{q}}_{t}$ and $\boldsymbol{\dot{q}}_{t-1}$ are the joint velocity vectors at the current and previous time steps, and $\Delta t$ is the simulation time step. Since MCRF penalizes sudden changes in motion, we expect ECIM policies to show lower joint acceleration than the PPO baseline.

\textbf{Joint torque (\({\tau}_t\))}: Joint torque represents the actuator effort at each joint during motion. It is an important indicator for both the naturalness of movement and the efficiency of energy use. We collect the joint torques at time step $t$ in the vector
\begin{equation}
    \boldsymbol{\tau}_t = [\tau_{1,t}, \tau_{2,t}, \ldots, \tau_{n,t}]^{T},
\end{equation}
where $\tau_{i,t}$ is the torque of the $i$-th joint at time step $t$, and $n$ is the number of actuated joints of the robot. Torque magnitude serves as a proxy for control intensity: higher values frequently indicate rapid actuator responses or inefficient force application, whereas lower values are associated with smoother motion profiles and greater energetic economy.

\textbf{Steps-to-$R^*$}: Steps-to-$R^*$ measures the number of environment interaction steps required for the agent to first achieve a target performance threshold $R^*$ (e.g., 90\% of the optimal policy’s average return). We compute it as
\begin{align}
\text{Steps-to-R*} = \min \left\{ N \,\middle|\, \frac{1}{K}\sum_{i=N-K+1}^{N} R_i \geq R^* \right\},
\end{align}
where $R_i$ is the cumulative reward of the $i$-th training episode, $K$ is the smoothing window size, and $R^*$ is the target performance threshold.

\textbf{Pitch RMS}: The root mean square of the body pitch angle (Pitch RMS) quantifies longitudinal trunk oscillations during locomotion. We compute it as
\begin{align}
\text{Pitch RMS} = \sqrt{\frac{1}{T} \sum_{t=1}^{T} \theta_t^2},
\end{align}
where $\theta_t$ is the pitch angle of the body relative to the horizontal plane at time $t$, and $T$ is the test time duration. 

\textbf{Acc RMS}: The root mean square of the body’s linear acceleration (Acc RMS) quantifies motion smoothness by reflecting dynamic impacts and high-frequency vibrations during locomotion. It is computed as
\begin{align}
\text{Acc RMS} = \sqrt{\frac{1}{T} \sum_{t=1}^{T} \|\mathbf{a}_t\|^2},
\end{align}
where $\mathbf{a}_t$ is the linear acceleration vector of the body at time $t$, and $T$ is the test duration. As with the other stability measures, MCRF is expected to reduce Acc RMS by promoting temporally consistent motion.

\textbf{Torque RMS}: The root mean square of joint torques (Torque RMS) represents the average magnitude of actuator torques across all joints, averaged over both time and joint index. It reflects energy consumption and actuation load. A lower value indicates more energy-efficient control, which is important for extending battery life and reducing hardware wear. We define it as
\begin{align}
\text{Torque RMS} = \sqrt{\frac{1}{T \cdot J} \sum_{t=1}^{T} \sum_{j=1}^{J} \tau_{t,j}^2},
\end{align}
where $\tau_{t,j}$ is the torque of joint $j$ at time $t$, $J$ is the number of joints, and $T$ is the test duration. This metric is mainly used to quantify the effect of IMDEEM on energy-related behavior.

\textbf{Attribution Gain (AG)}: To quantify the contribution of each ECIM module in the ablation study, we define Attribution Gain as
\begin{align}
\text{AG} = \frac{1}{E} \sum_{e=1}^{E} \left( M_{\text{full},e} - M_{\text{minus-}m,e} \right),
\end{align}
where $ M $ denotes a specific performance metric, like Reward and Steps-to-R*, $ e $ ranges over all $ E=6 $ environments, and $ m $ represents the ablated module. A higher AG value indicates a greater influence of the module to that metric.

\subsection{ECIM vs PPO: Entropy-Controlled Intrinsic Motivation in Complex Terrains}

To comprehensively evaluate the proposed ECIM algorithm as a whole framework, we first compare it against the PPO baseline across the six terrain types across varying complexity as well: flat, sloped, rough, stair-climbing, stair-descending, and stepping-stones.

\subsubsection{Learning Efficiency and Task Reward under Entropy-Controlled Exploration}

\begin{figure}[htbp]
\centering

\subfloat[Reward in Flat Terrain]{%
  \resizebox*{4cm}{!}{\includegraphics{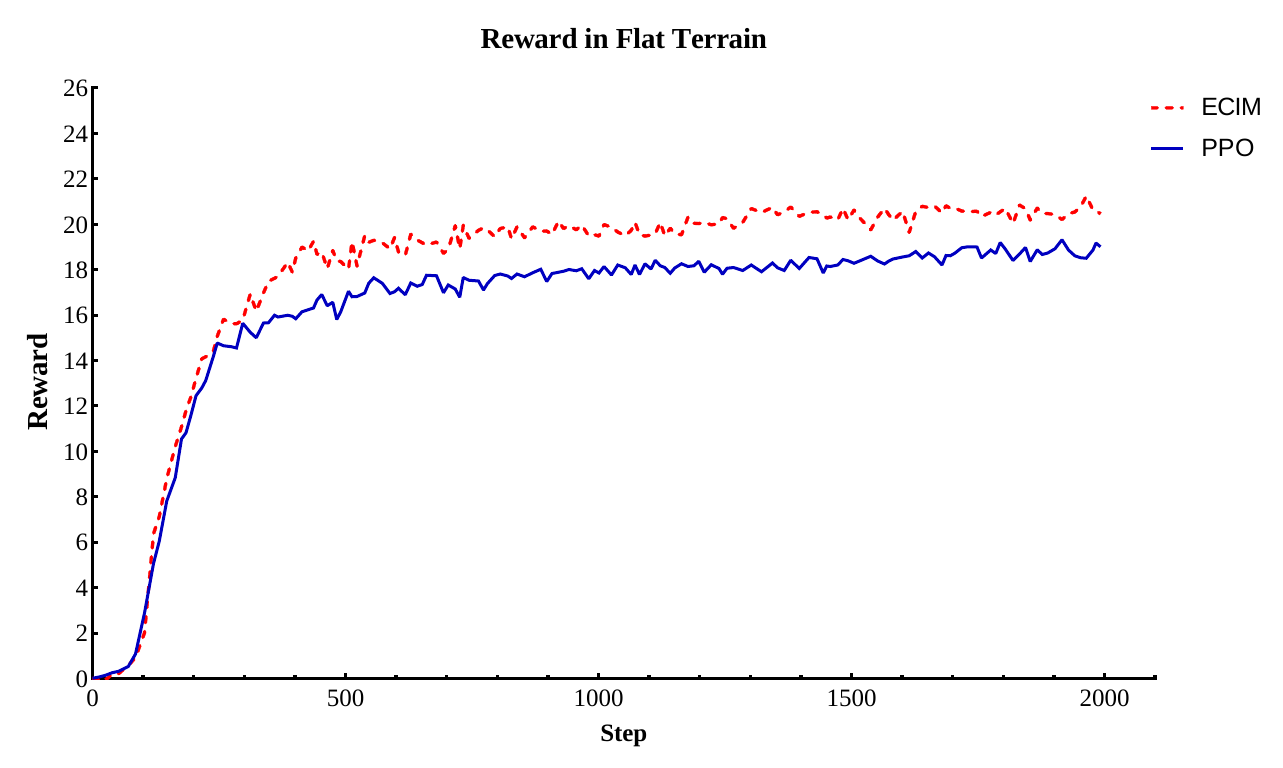}}}\hspace{5pt}
\subfloat[Reward in Sloped Terrain]{%
  \resizebox*{4cm}{!}{\includegraphics{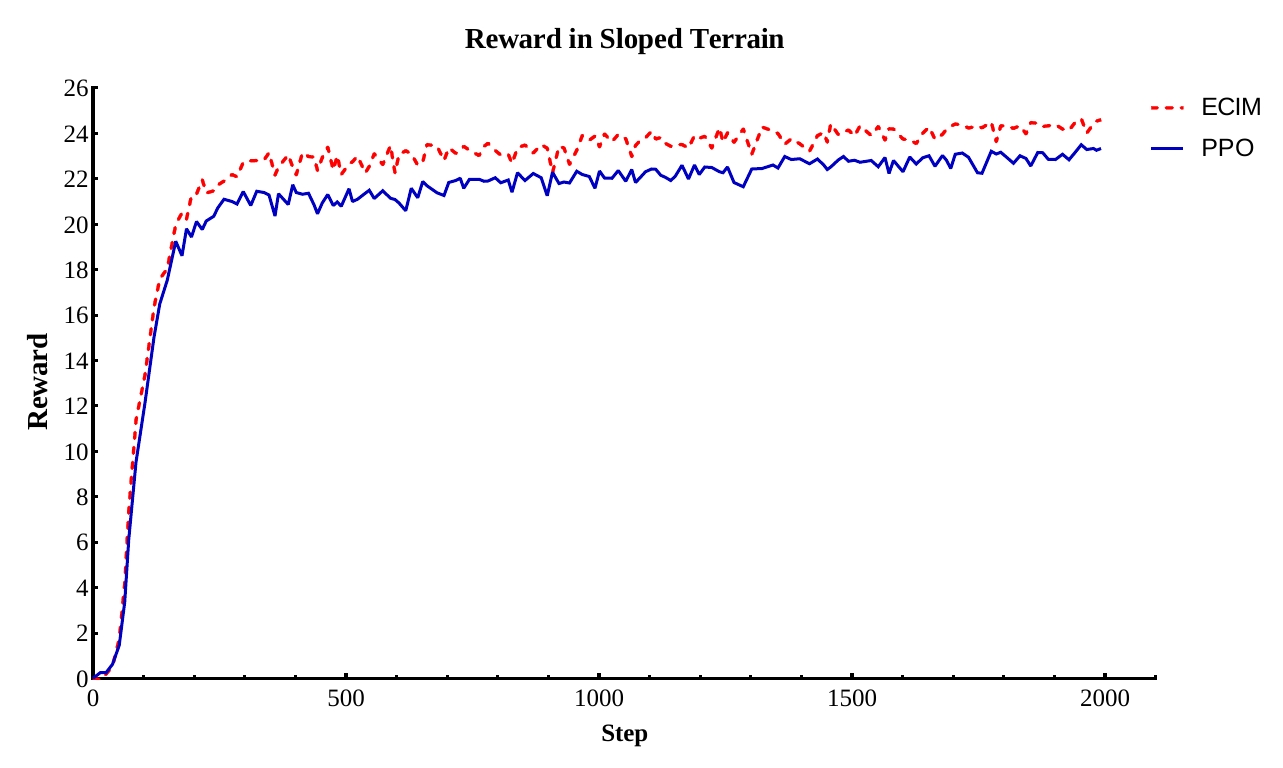}}}\hspace{5pt}
\subfloat[Reward in Rough Terrain]{%
  \resizebox*{4cm}{!}{\includegraphics{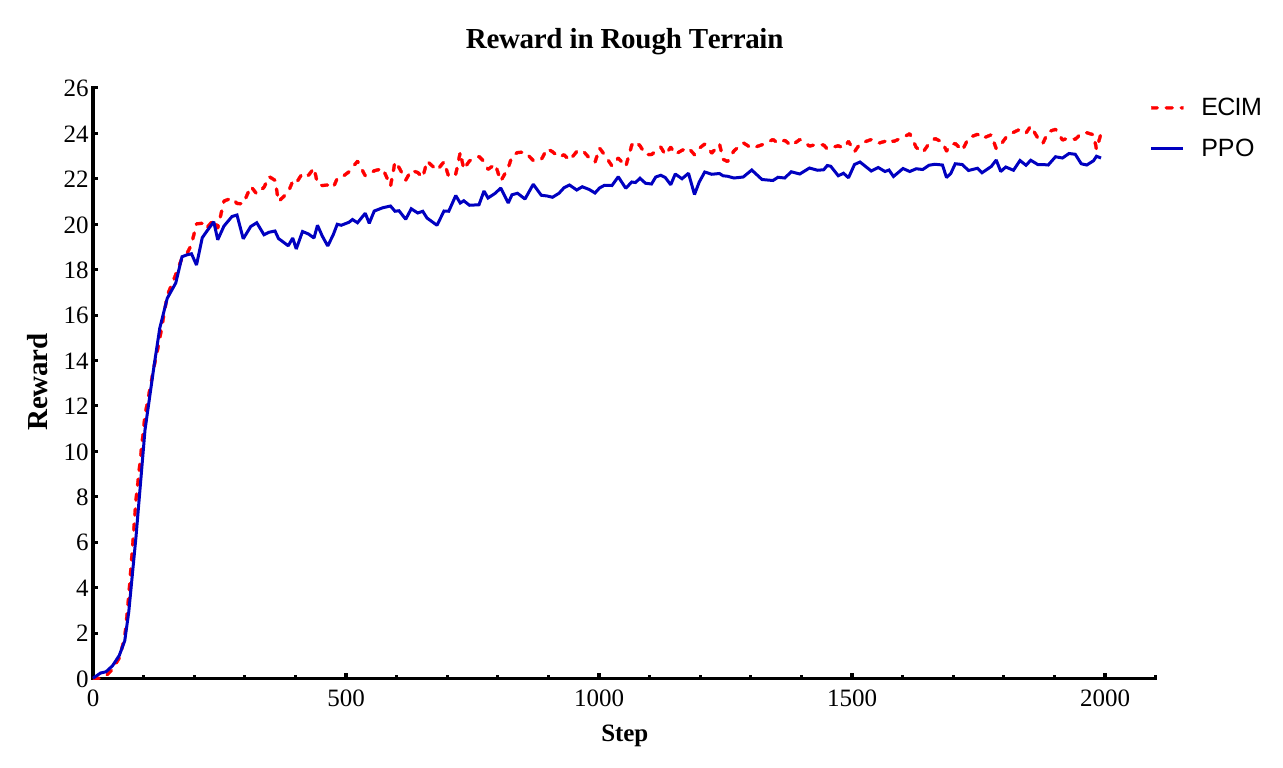}}}\hspace{5pt}
  
\subfloat[Reward in Stair-Climbing Terrain]{%
  \resizebox*{3.9cm}{!}{\includegraphics{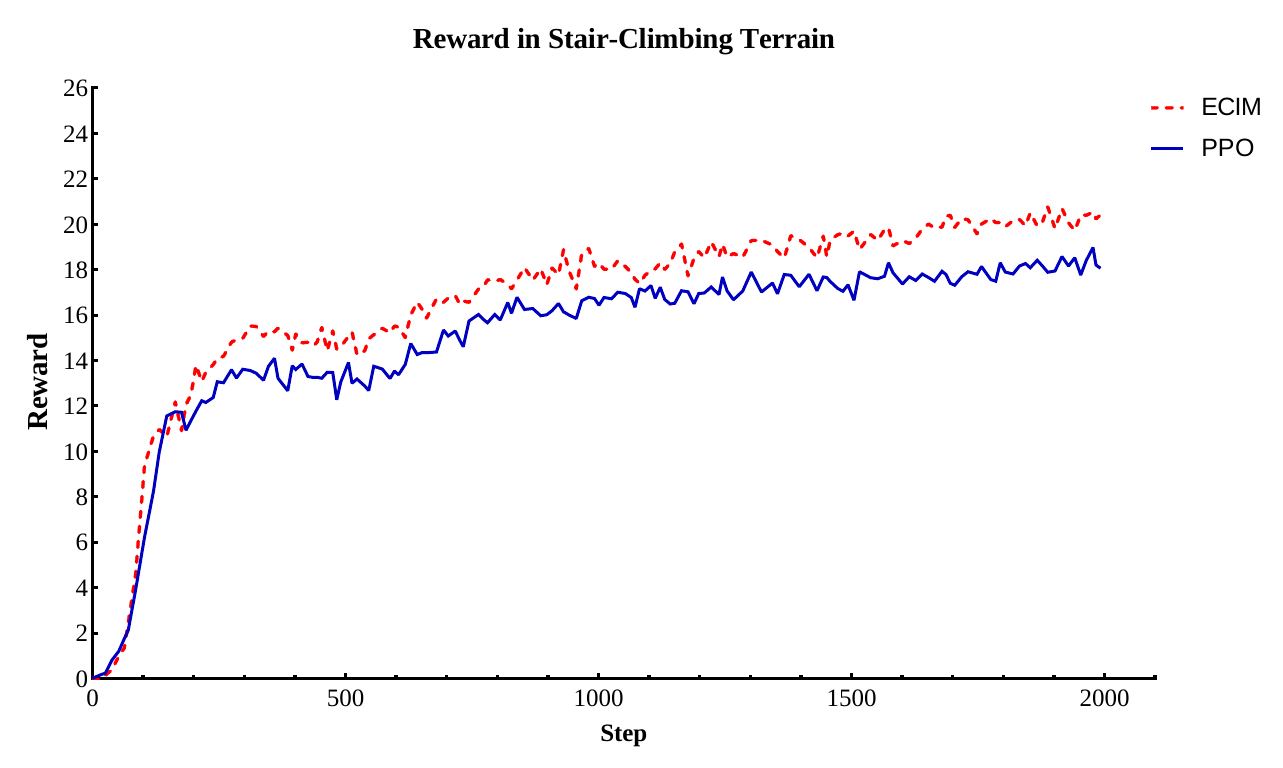}}}\hspace{5pt}
\subfloat[Reward in Stair-Descending Terrain]{%
  \resizebox*{4cm}{!}{\includegraphics{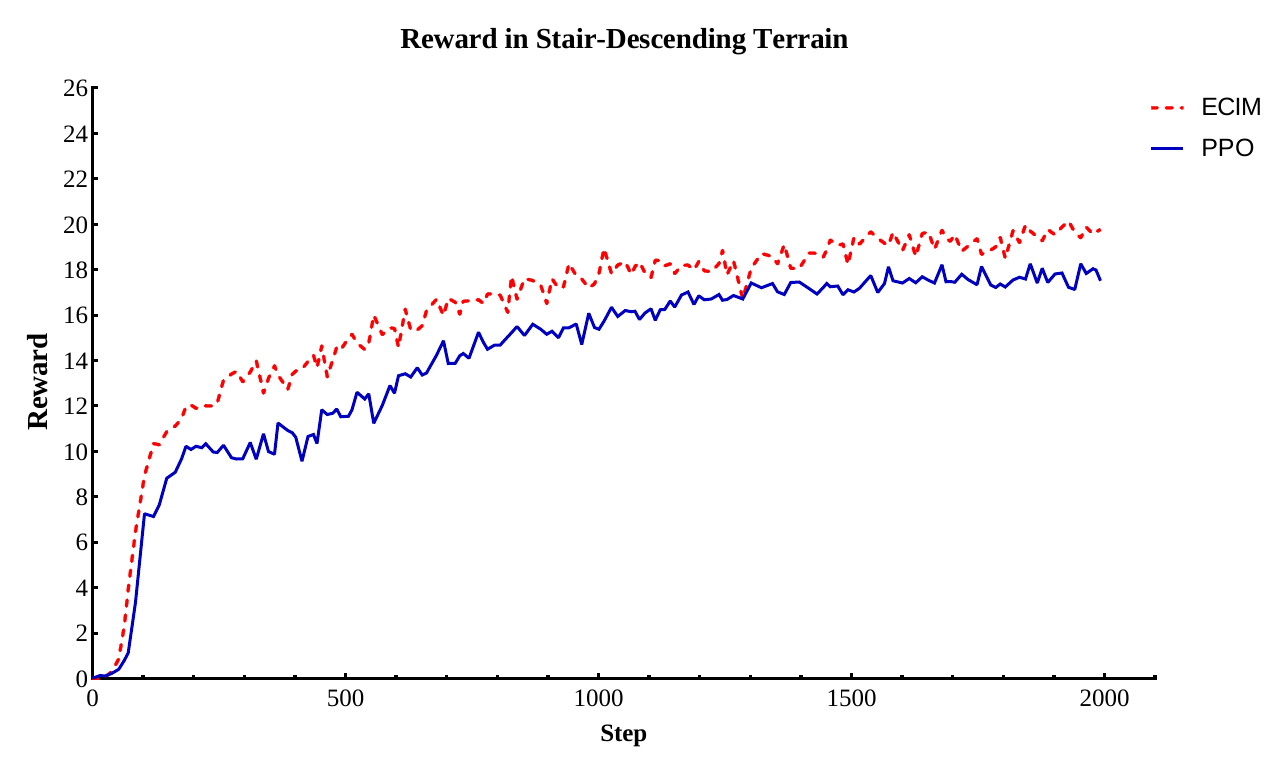}}}\hspace{5pt}
\subfloat[Reward in Stepping Stones Terrain]{%
  \resizebox*{4cm}{!}{\includegraphics{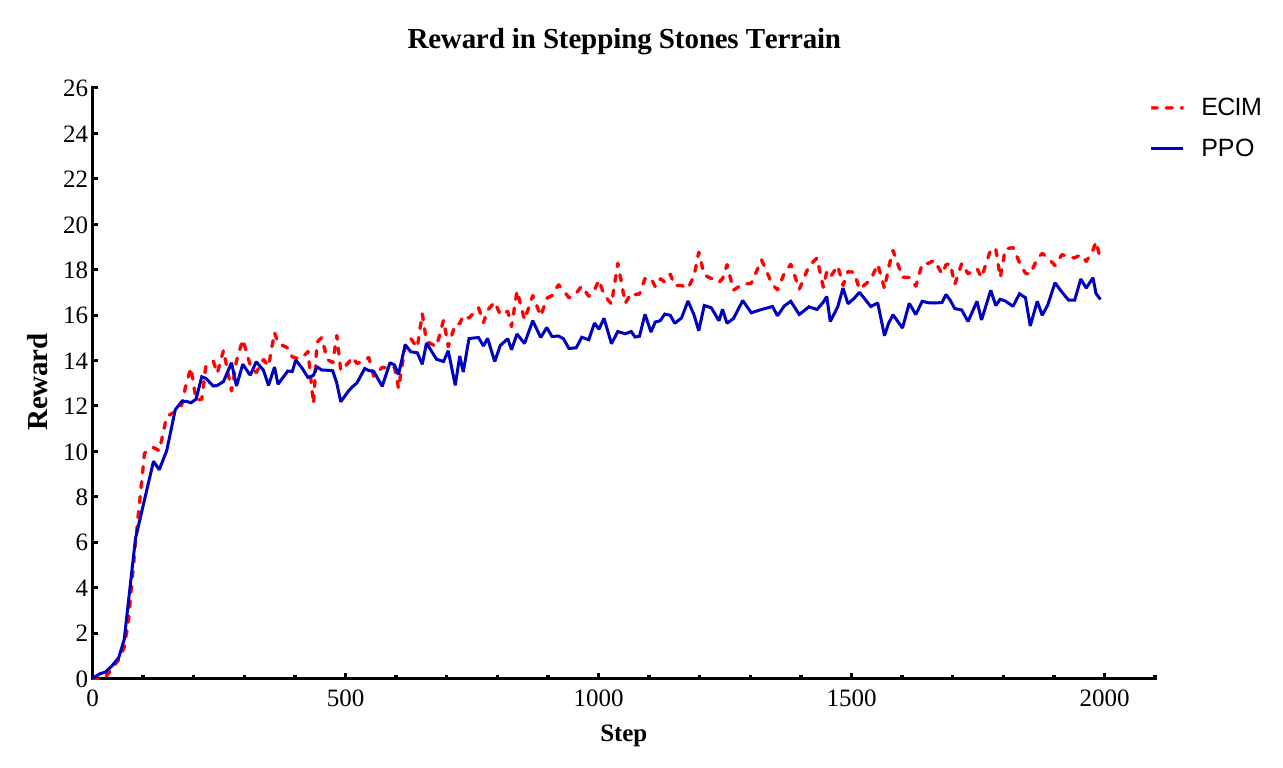}}}\hspace{5pt}
\caption{Robot average reward variations under six different terrain environments for PPO and ECIM.}
\label{fig:reward}
\end{figure}
                                                      
\begin{figure}[htbp]
\centering
\subfloat[Reward in Flat Terrain]{%
  \resizebox*{4cm}{!}{\includegraphics{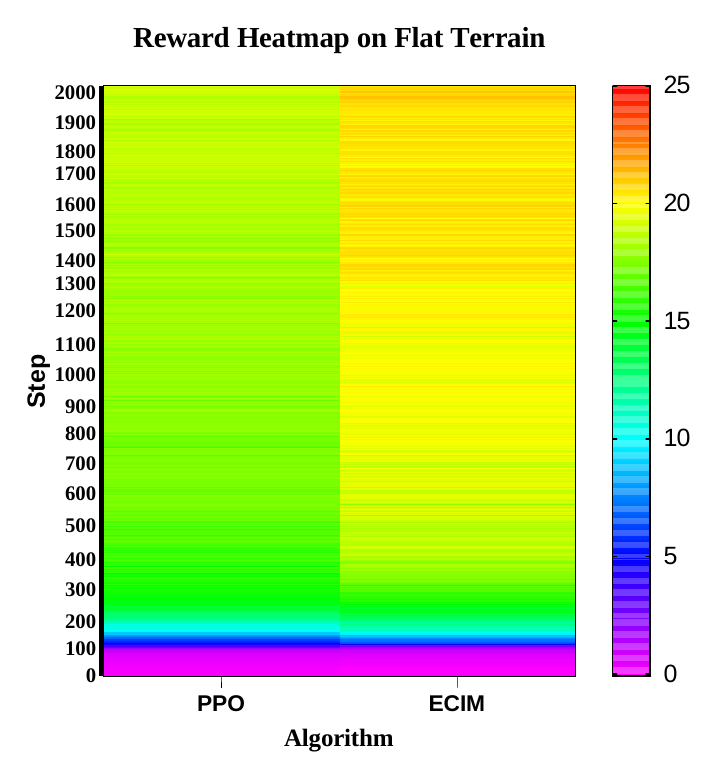}}}\hspace{5pt}
\subfloat[Reward in Sloped Terrain]{%
  \resizebox*{4cm}{!}{\includegraphics{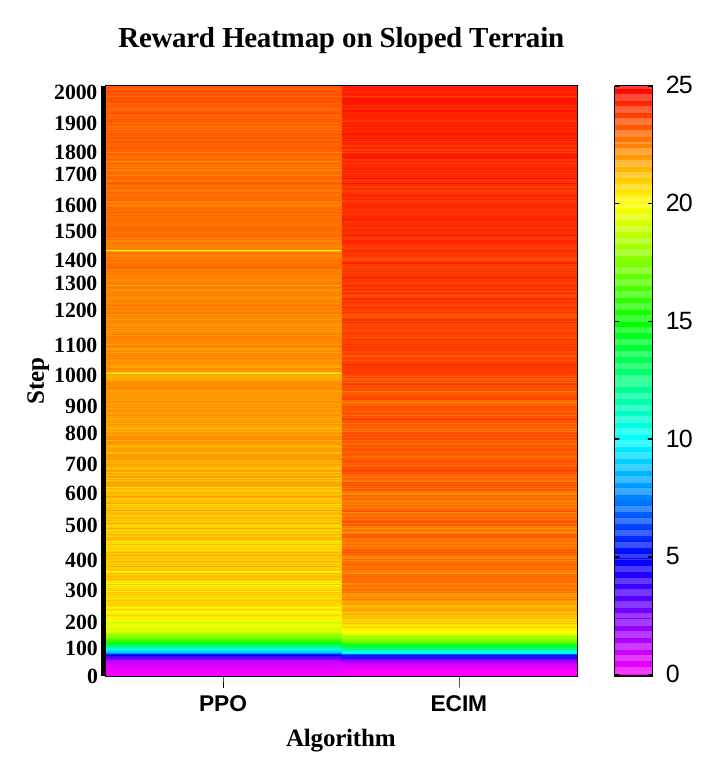}}}\hspace{5pt}
\subfloat[Reward in Rough Terrain]{%
  \resizebox*{4cm}{!}{\includegraphics{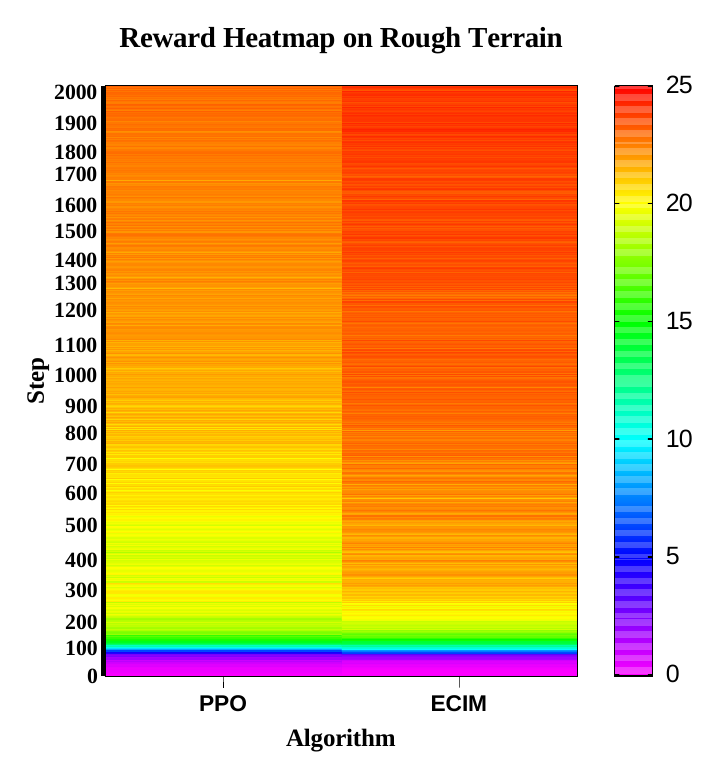}}}\hspace{5pt}
\subfloat[Reward in Stair-Climbing Terrain]{%
  \resizebox*{4cm}{!}{\includegraphics{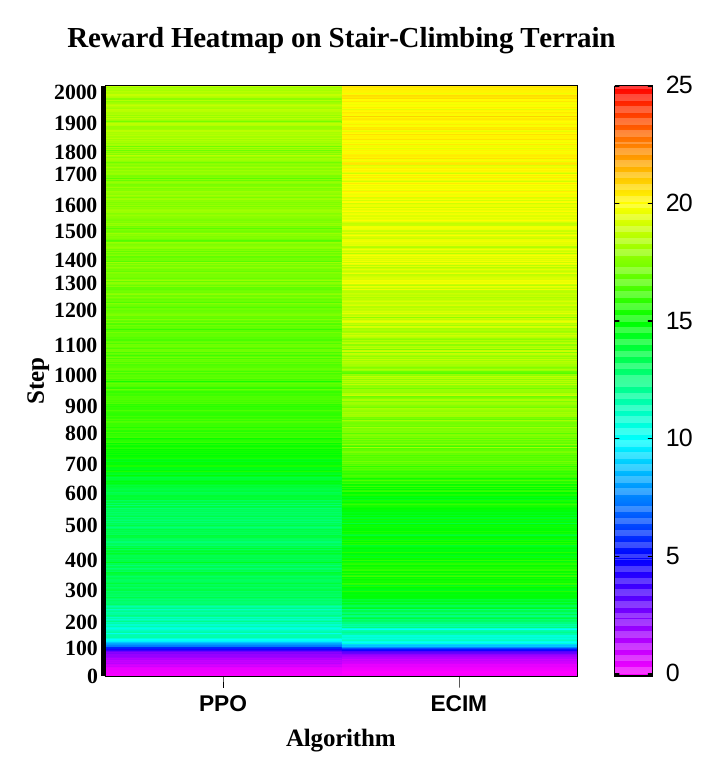}}}\hspace{5pt}
\subfloat[Reward in Stair-Descending Terrain]{%
  \resizebox*{4cm}{!}{\includegraphics{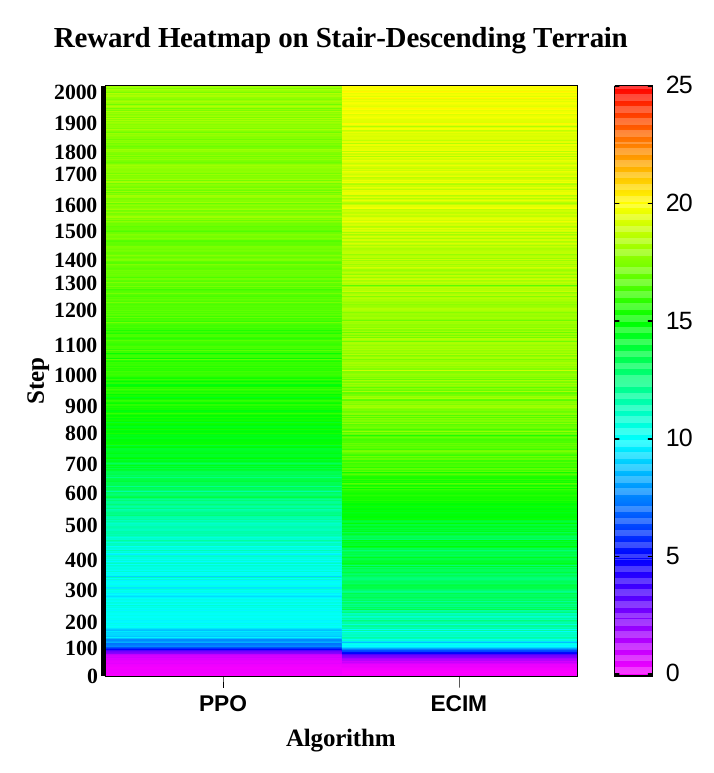}}}\hspace{5pt}
\subfloat[Reward in Stepping Stones Terrain]{%
  \resizebox*{4cm}{!}{\includegraphics{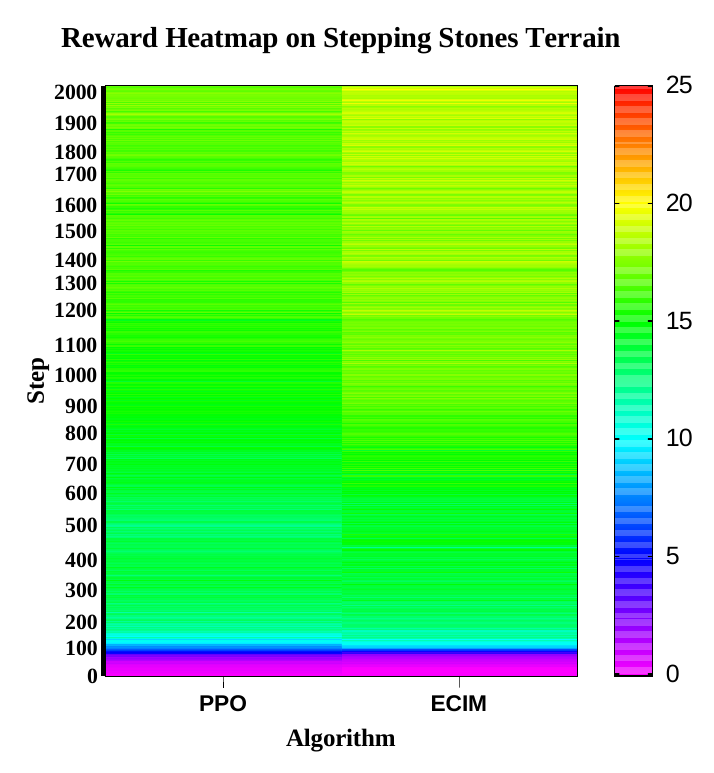}}}
\caption{Comparison of reward heatmaps for the PPO and ECIM algorithms across different terrains. The vertical axis represents training steps, with color intensity reflecting the magnitude of rewards obtained by the two algorithms at corresponding steps: different colors denote the reward levels of the ECIM and PPO algorithms, respectively.
}
\label{fig:reward heatmaps}
\end{figure}

The evolution of cumulative rewards is the main indicator we use to assess learning efficiency and task performance in our setting. It reflects how quickly the robot learns a useful policy and how well it can stabilize and coordinate its motion on each terrain. Figure~\ref{fig:reward} shows the average reward curves of PPO and ECIM across the six terrain environments. Figure~\ref{fig:reward heatmaps} further visualizes the reward evolution as heatmaps over training steps, making it easier to compare learning speed and final performance for the two algorithms.

At the beginning of training, PPO and ECIM show similar reward growth and reach basic locomotion skills at roughly the same pace. This shows that ECIM keeps the strong exploration ability of original PPO and can quickly learn fundamental gait patterns. As training continues, however, the differences between the two methods become more obvious, especially on the more complex terrains. On these terrains, PPO’s reward curves tend to grow more slowly and enter a plain earlier, which suggests that the policy may stuck in local optimal values. In contrast, ECIM maintains a higher learning rate in this critical phase, and the rewards continue to increase over a longer period. This behaviour shows consistent output with the design of our pipeline, where AECPOM and IMDEEM work together to adapt exploration and provide additional learning signals beyond the standard PPO objective to give a bigger picture hint of exploration for the model.

The Table~\ref{tab:performance_summary} summarizes the quantitative improvements of ECIM over benchmarks across all mentioned metrics and terrain complexity levels. For cumulative reward, ECIM achieves approximately $+10.5\%$ improvement on the simple terrain (flat), $+4.3\%$ on the medium-complexity terrains (sloped, rough), and $+11.1\%$--$11.8\%$ on the complex terrains (stairs and stepping-stones), leading to an overall gain of $+8.6\%$. At the same time, Pitch Angle RMS and Joint Acceleration RMS are reduced by around $-25\%$ and $-27\%$ on average, and Torque RMS is reduced by about $-16\%$. These results show that the full ECIM pipeline not only improves reward, but also produces more stable and smoother motion with lower torque usage. In combination, the reward curves, heatmaps and aggregated metrics confirm that ECIM supports faster and more reliable policy learning than PPO, and that the benefits are most pronounced on the more challenging terrains.

\begin{table}[htbp]
\centering
\caption{Performance improvements of ECIM over PPO across six terrains.}
\label{tab:performance_summary}
\begin{tabular}{lcccc}
\toprule
\textbf{Metric} & \textbf{Simple} & \textbf{Medium} & \textbf{Complex} & \textbf{Overall} \\
\midrule
Cumulative Reward (\%) & +10.5 & +4.3 & +11.1--11.8 & +8.6 \\
Pitch Angle RMS (\%) & -25.2 & -23.1 & -27.6--29.4 & -26.3 \\
Joint Acceleration RMS (\%) & -20.3 & -24.7 & -28.9--32.4 & -26.6 \\
Torque RMS (\%) & -11.3 & -14.8 & -17.2--20.0 & -15.8 \\
\bottomrule
\multicolumn{5}{l}{\small \textit{Simple: Flat; Medium: Sloped, Rough; Complex: Stairs Up/Down, Stepping Stones}}
\end{tabular}
\end{table}

\subsubsection{Locomotion Stability and Smoothness in Intrinsically Motivated Gaits}

\begin{figure}[htbp]
\centering
\subfloat[Pitch Angle in Flat Terrain]{%
  \resizebox*{4cm}{!}{\includegraphics{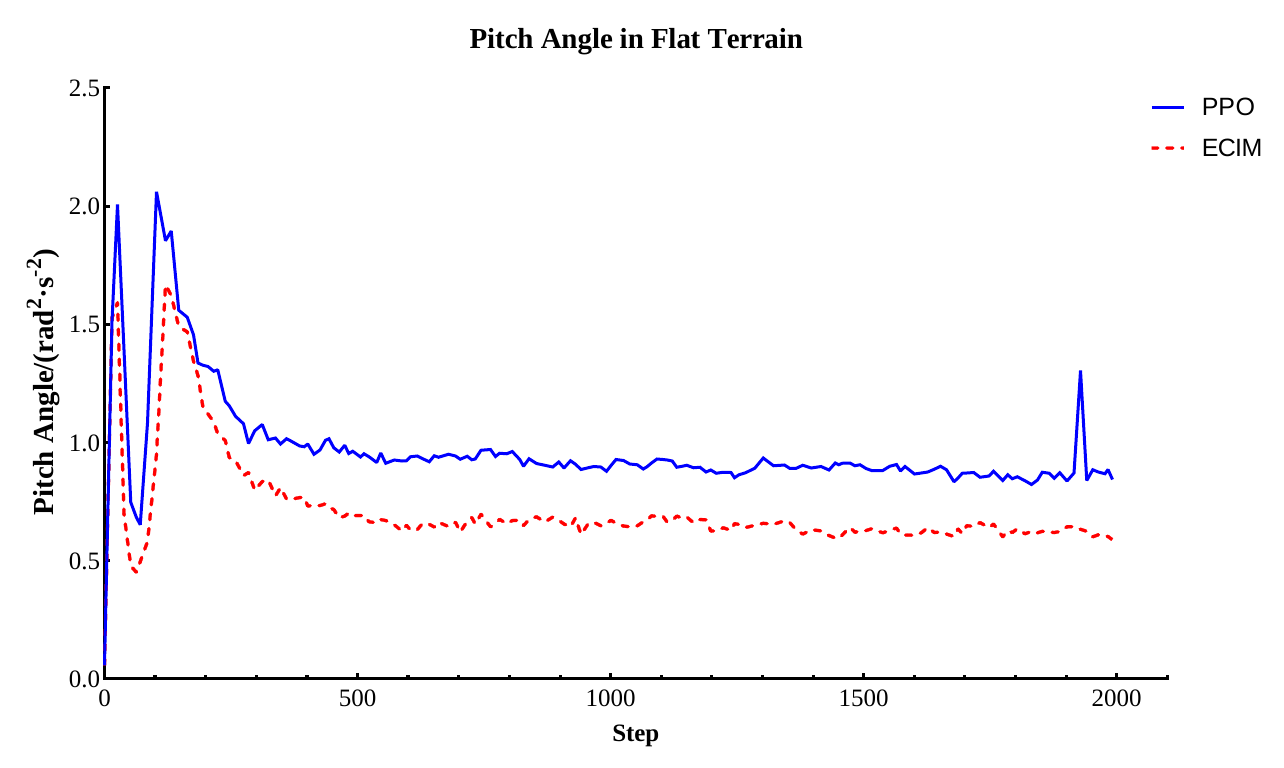}}}\hspace{5pt}
\subfloat[Pitch Angle in Sloped Terrain]{%
  \resizebox*{4cm}{!}{\includegraphics{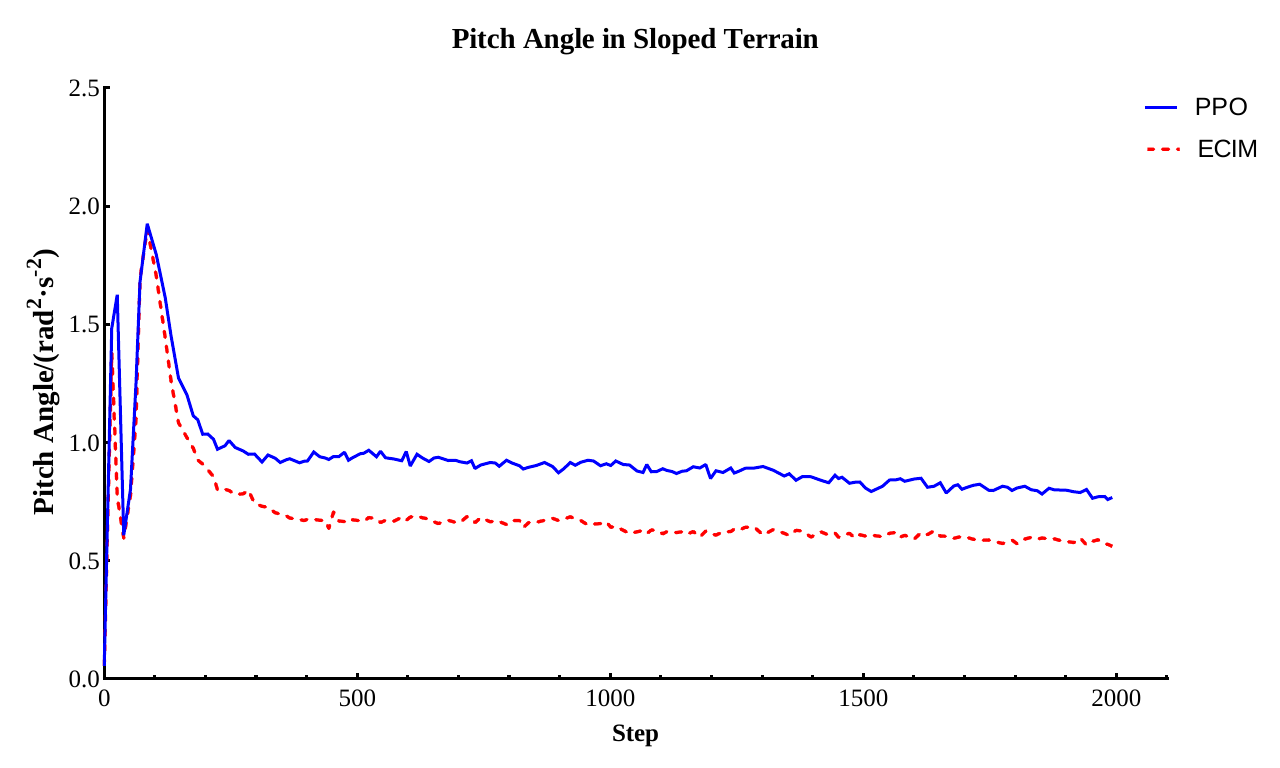}}}\hspace{5pt}
\subfloat[Pitch Angle in Rough Terrain]{%
  \resizebox*{4cm}{!}{\includegraphics{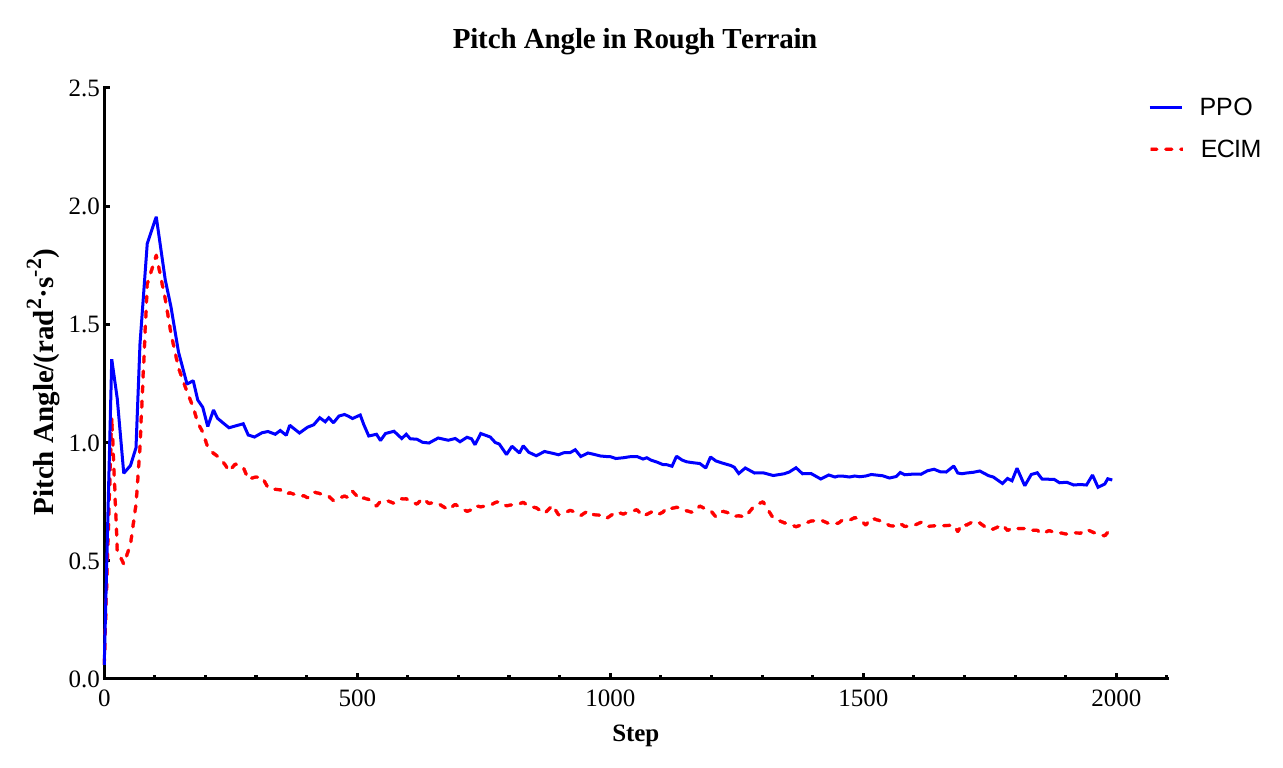}}}\hspace{5pt}
\subfloat[Pitch Angle in Stair-Climbing Terrain]{%
  \resizebox*{4cm}{!}{\includegraphics{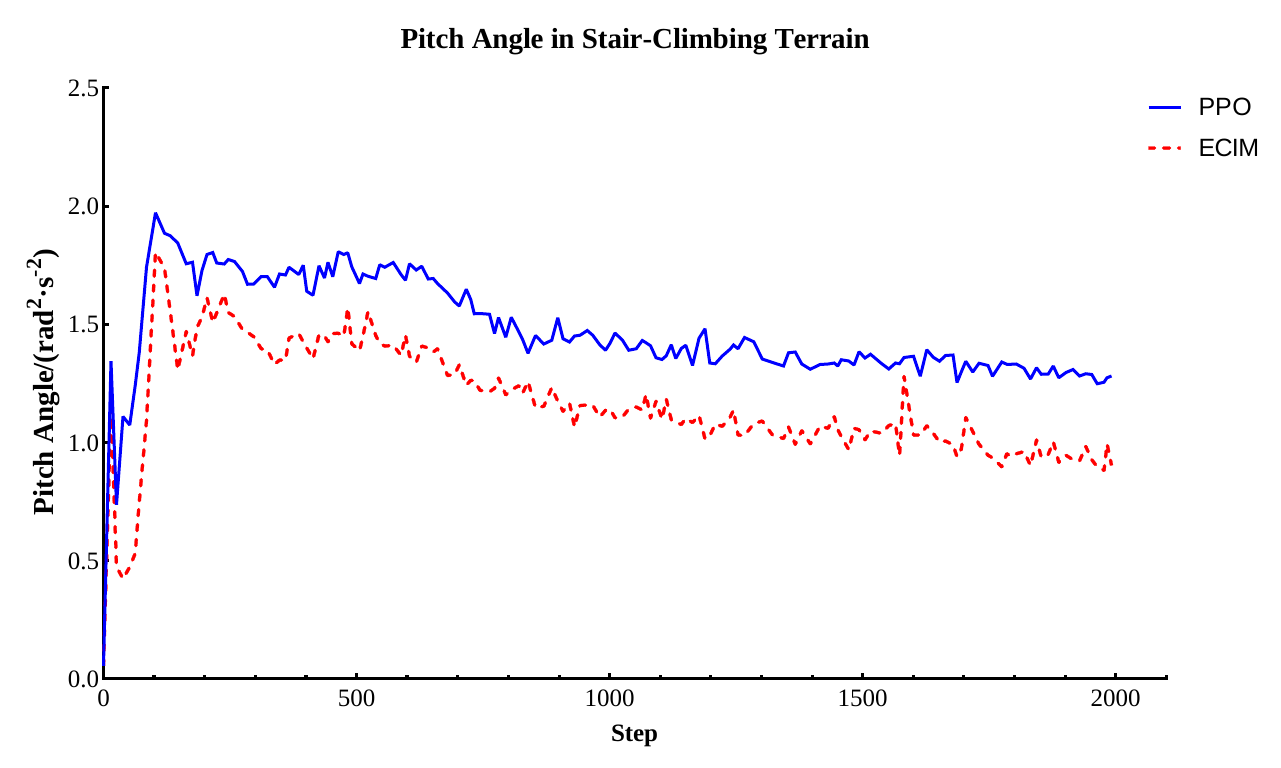}}}\hspace{5pt}
\subfloat[Pitch Angle in Stair-Descending Terrain]{%
  \resizebox*{4cm}{!}{\includegraphics{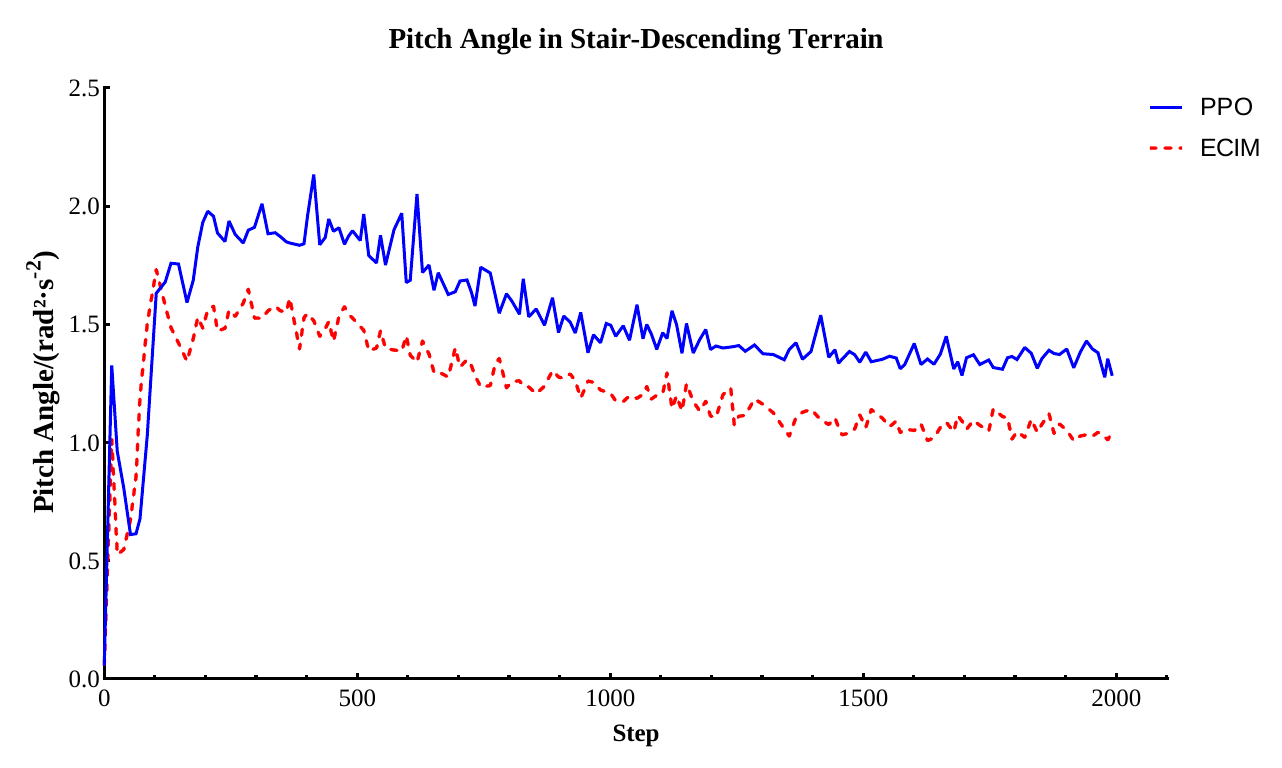}}}\hspace{5pt}
\subfloat[Pitch Angle in Stepping Stones Terrain]{%
  \resizebox*{4cm}{!}{\includegraphics{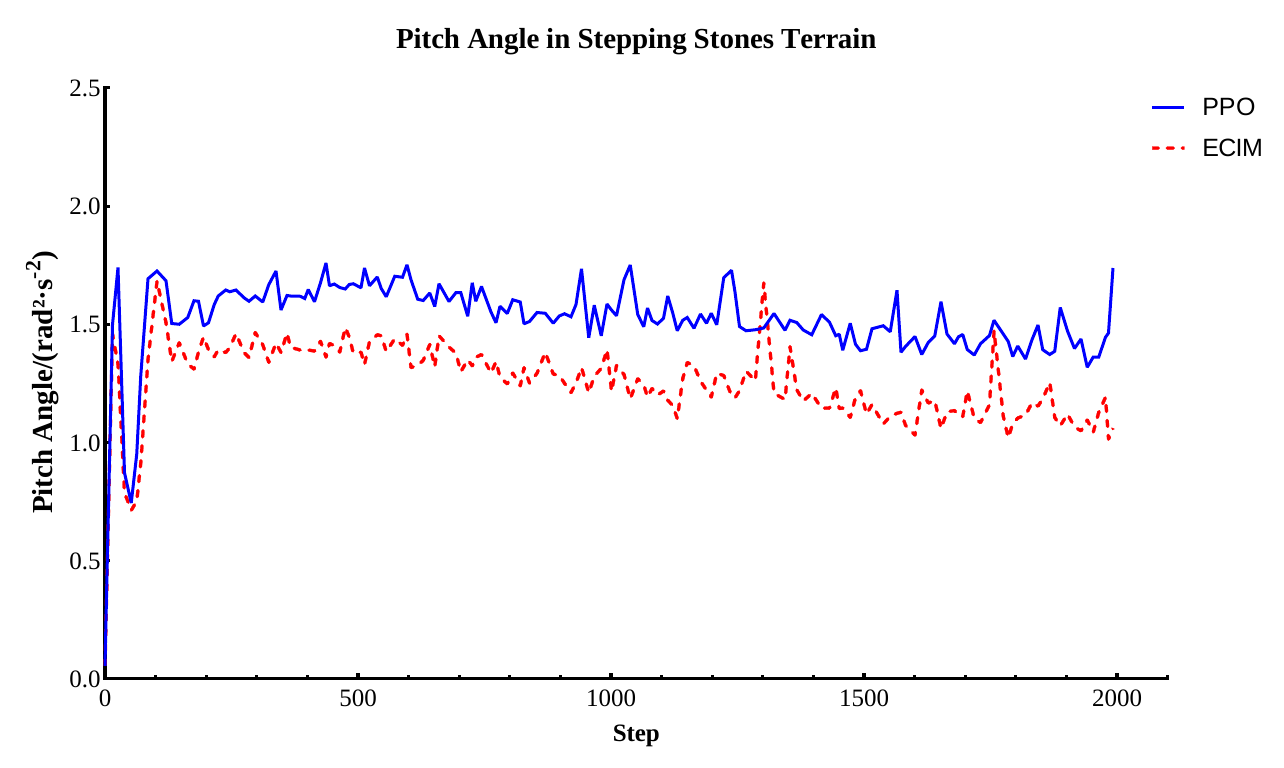}}}\hspace{5pt}
\caption{Pitch angle variations under six different terrain environments for PPO and ECIM.}
\label{fig:ang}
\end{figure}

\begin{figure}[htbp]
\centering
\subfloat[Joint Acceleration in Flat Terrain]{%
  \resizebox*{4cm}{!}{\includegraphics{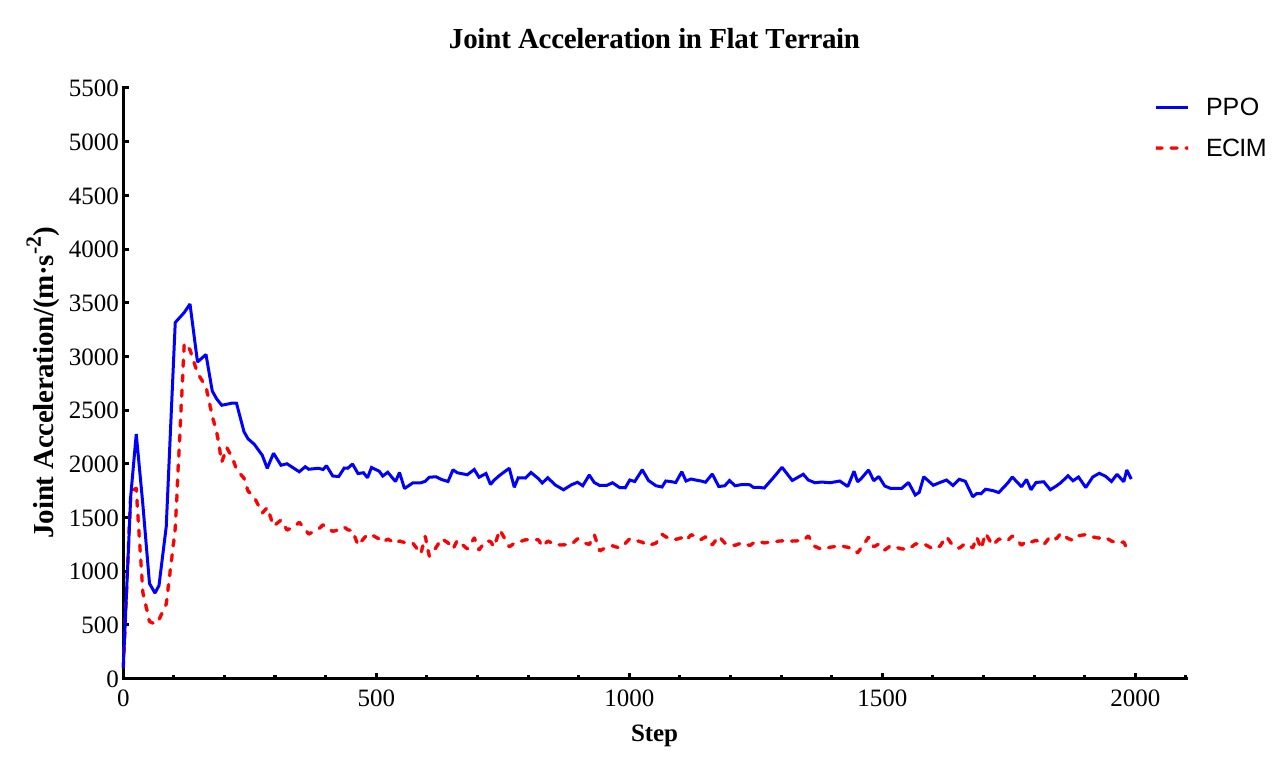}}}\hspace{5pt}
\subfloat[Joint Acceleration in Sloped Terrain]{%
  \resizebox*{4cm}{!}{\includegraphics{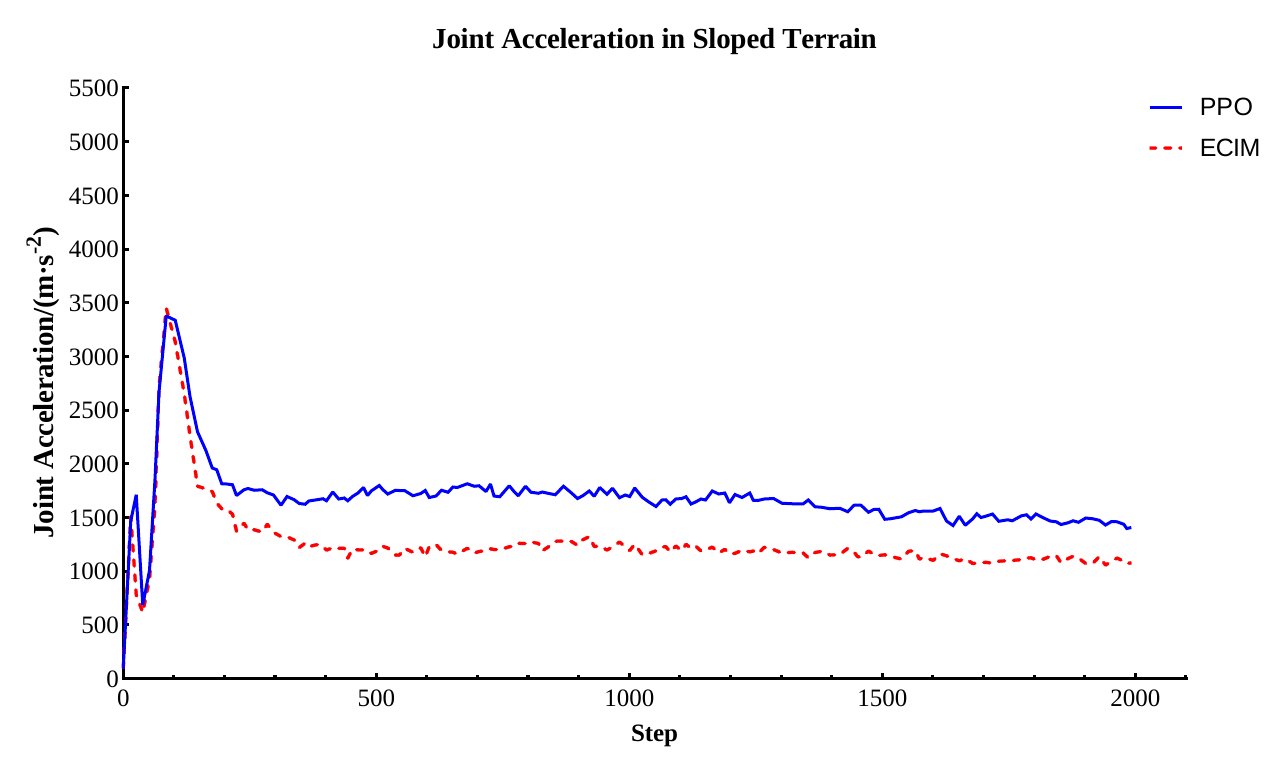}}}\hspace{5pt}
\subfloat[Joint Acceleration in Rough Terrain]{%
  \resizebox*{4cm}{!}{\includegraphics{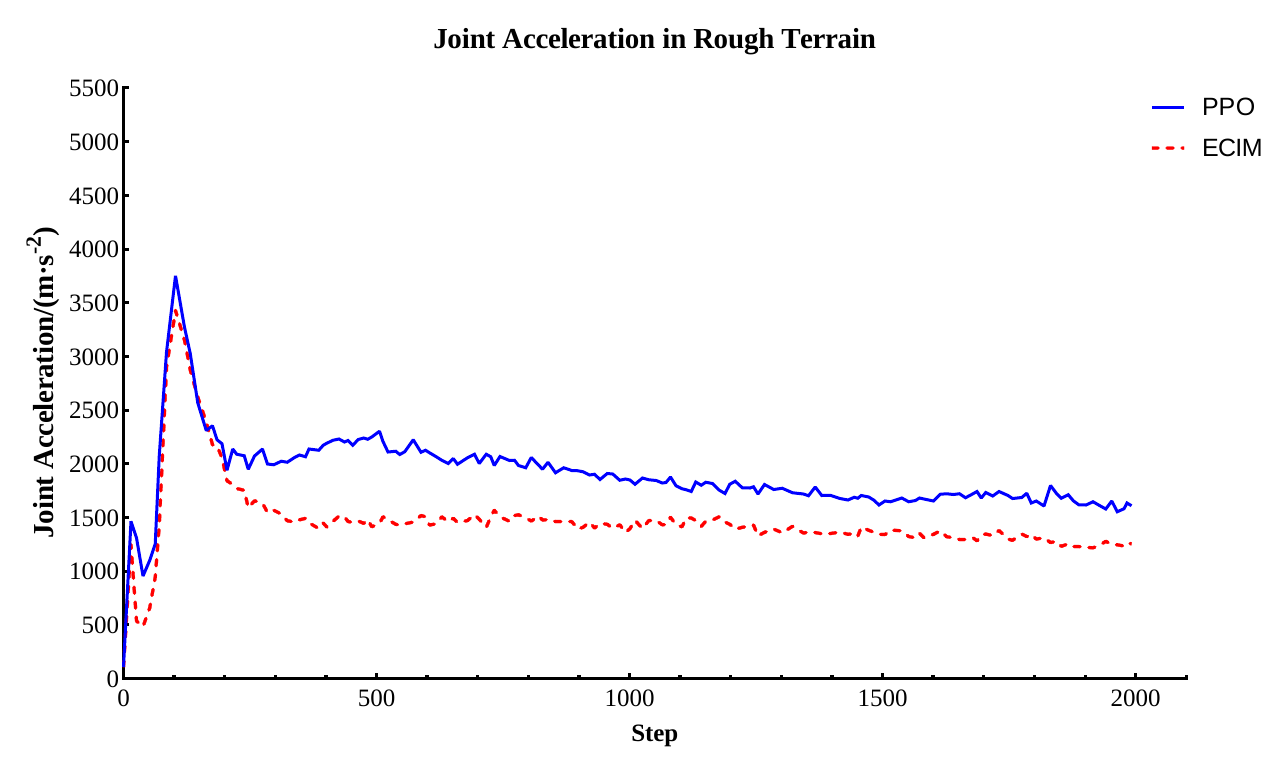}}}\hspace{5pt}
\subfloat[Joint Acceleration in Stair-Climbing Terrain]{%
  \resizebox*{4cm}{!}{\includegraphics{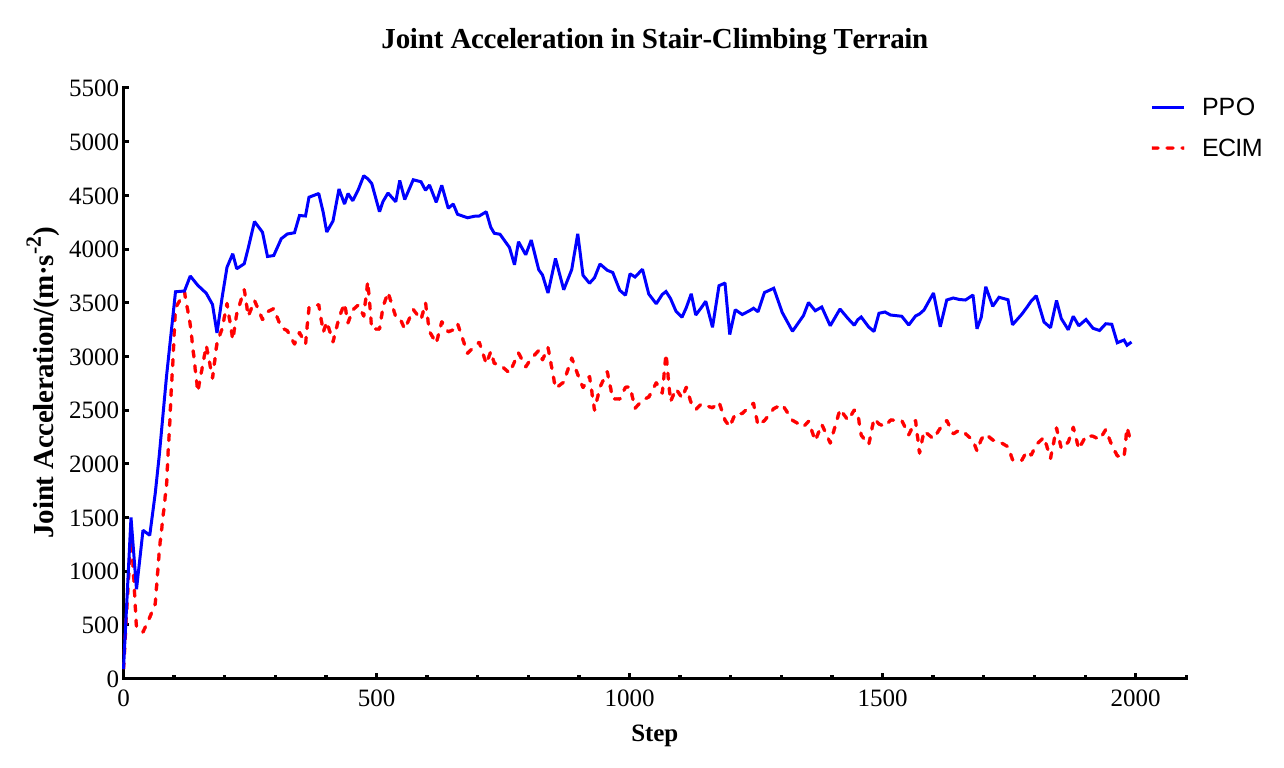}}}\hspace{5pt}
\subfloat[Joint Acceleration in Stair-Descending Terrain]{%
  \resizebox*{4cm}{!}{\includegraphics{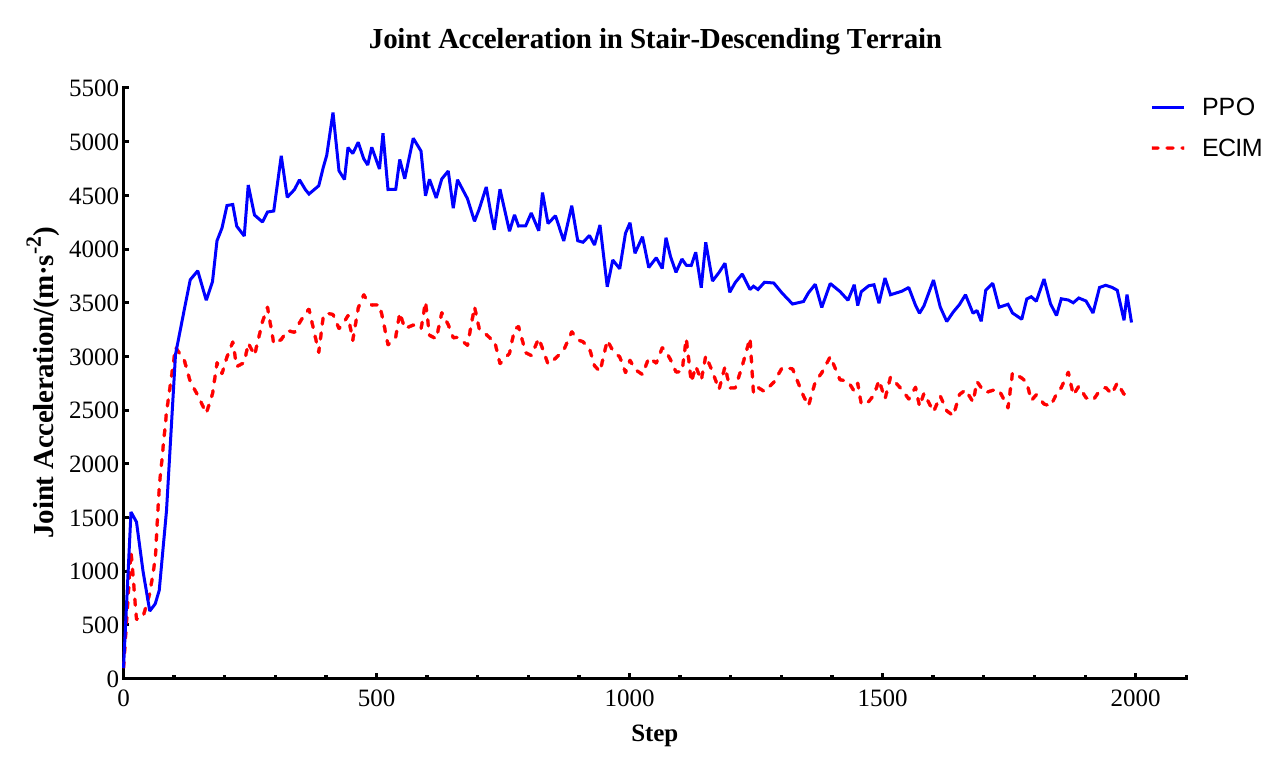}}}\hspace{5pt}
\subfloat[Joint Acceleration in Stepping Stones Terrain]{%
  \resizebox*{4cm}{!}{\includegraphics{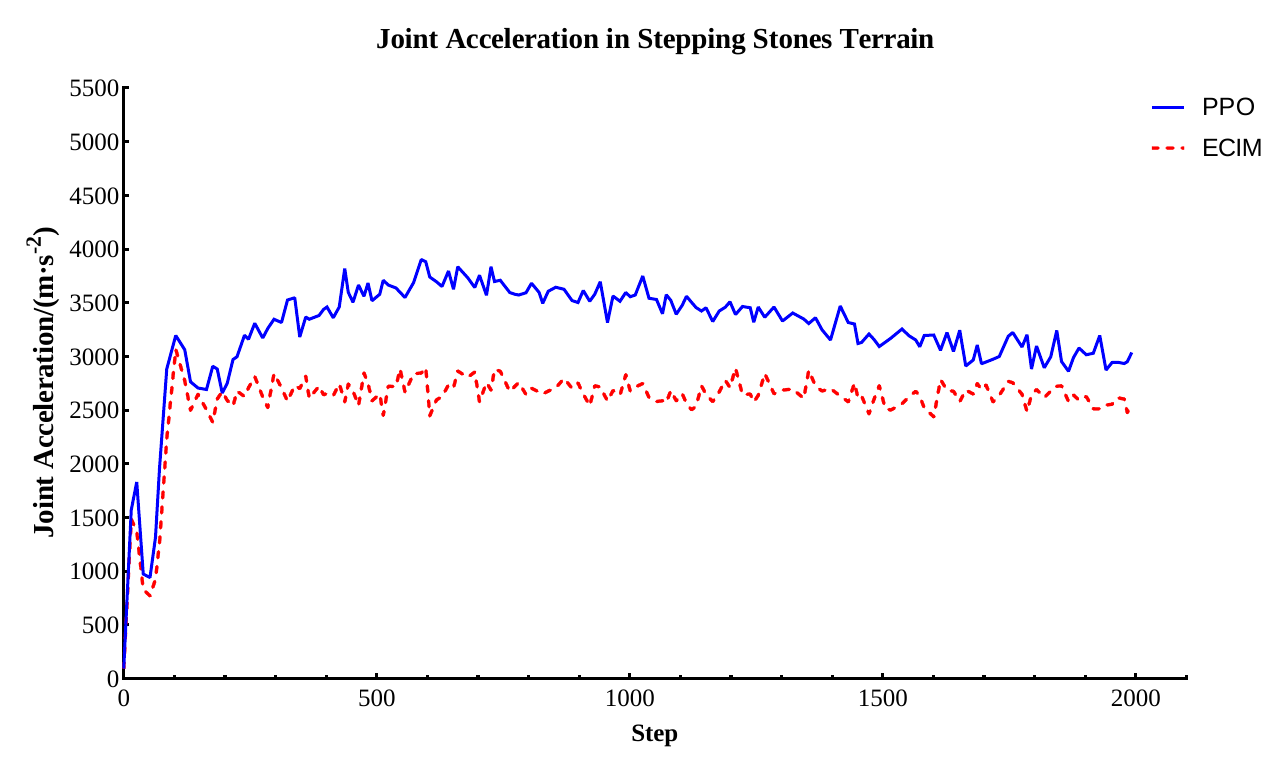}}}\hspace{5pt}
\caption{Joint acceleration variations under six different terrain environments for PPO and ECIM.}
\label{fig:acc}
\end{figure}

The above Figure~\ref{fig:ang} shows the pitch angle evolution of PPO and ECIM on all six terrains. Overall, ECIM improves posture stability by about $23.1\%$--$29.4\%$ compared with PPO (see Table~\ref{tab:performance_summary}). In all environments, both algorithms reduce the initially large pitch deviations and converge towards a stable range. However, the ECIM curves stay consistently below those of PPO, and converge to smaller mean pitch values with narrower fluctuation ranges. On the more complex terrains, such as stair climbing, stair descending and stepping stones, PPO shows larger oscillations and sometimes stagnation in the later training stages, while ECIM maintains smooth, low-amplitude pitch trajectories. This indicates that the proposed motion continuity regularization, together with the intrinsic reward design, effectively suppresses unstable body tilting and helps the robot keep a more stable trunk posture throughout training.

The joint acceleration results in Figure~\ref{fig:acc} further highlight the improvement in motion smoothness. Across all terrains, ECIM reduces the joint acceleration RMS by approximately $20.3\%$--$32.4\%$ relative to PPO (see Table~\ref{tab:performance_summary}). Both methods start with high joint accelerations and then quickly move into a more stable range during early training. In complex terrains, however, PPO often shows slower convergence or oscillations in joint acceleration curves, while ECIM continues to optimize and eventually converges to lower average acceleration with smooth, low-oscillation trajectories. These results confirm that the ECIM framework not only stabilizes the robot’s body attitude, but also produces smoother and more natural gait transitions. In addition, smoother accelerations are consistent with the observed reduction in torque usage, which shows that ECIM improves motion quality without relying on overly actuation effort.

\subsubsection{Energy Efficiency of Entropy-Controlled Quadruped Policies}

\begin{figure}[htbp]
\centering
\subfloat[Torque in Flat Terrain]{%
  \resizebox*{4cm}{!}{\includegraphics{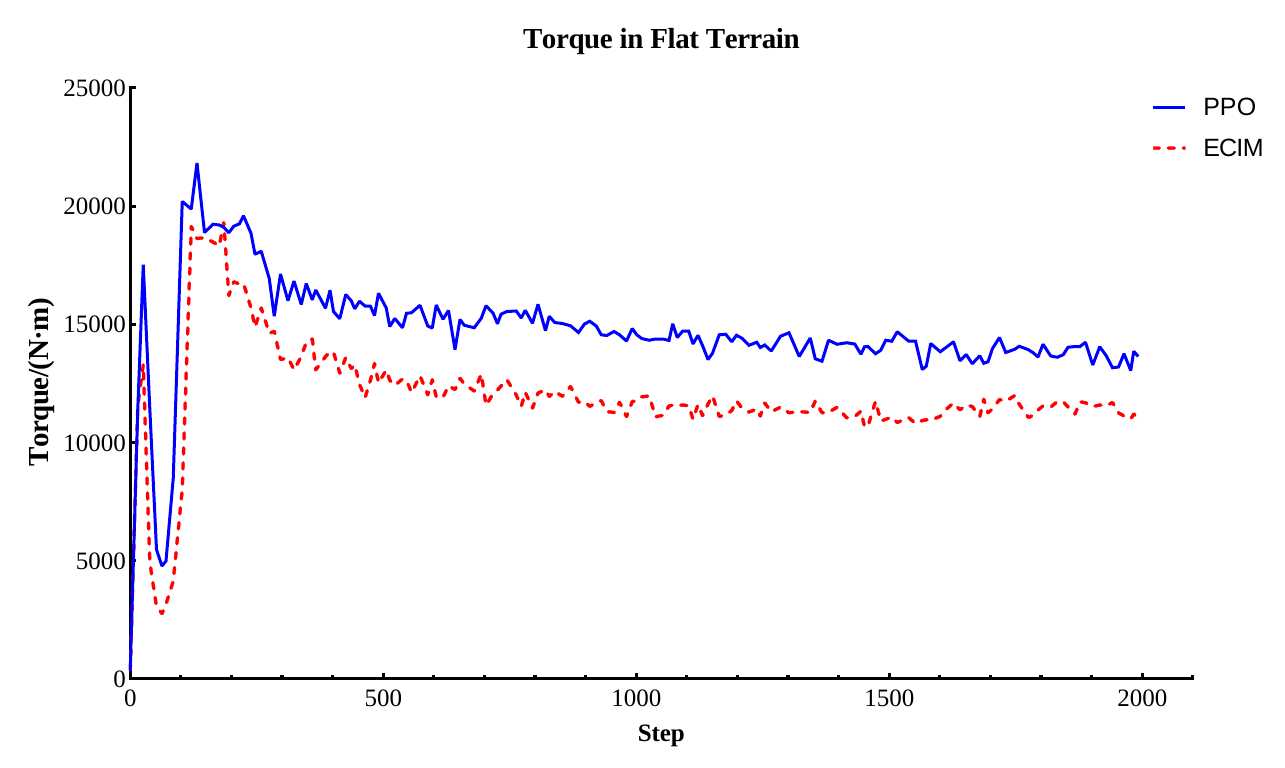}}}\hspace{5pt}
\subfloat[Torque in Sloped Terrain]{%
  \resizebox*{4cm}{!}{\includegraphics{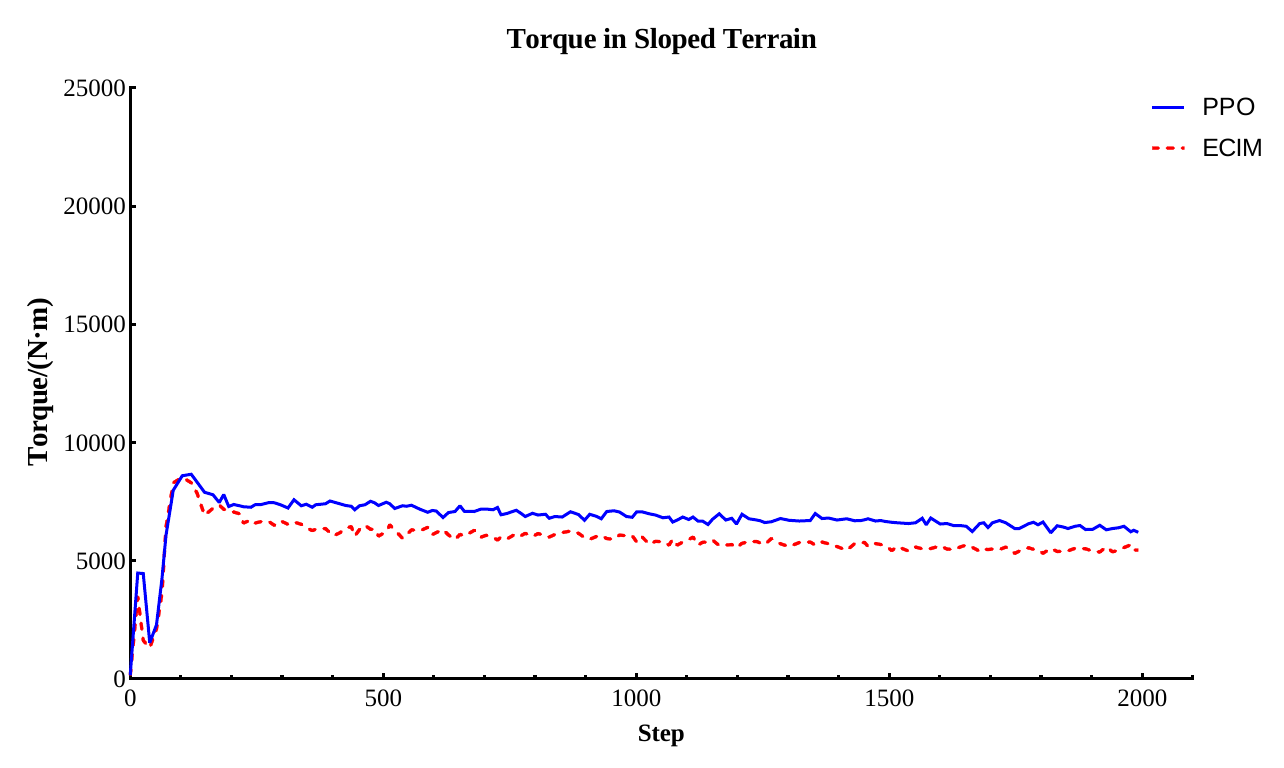}}}\hspace{5pt}
\subfloat[Torque in Rough Terrain]{%
  \resizebox*{4cm}{!}{\includegraphics{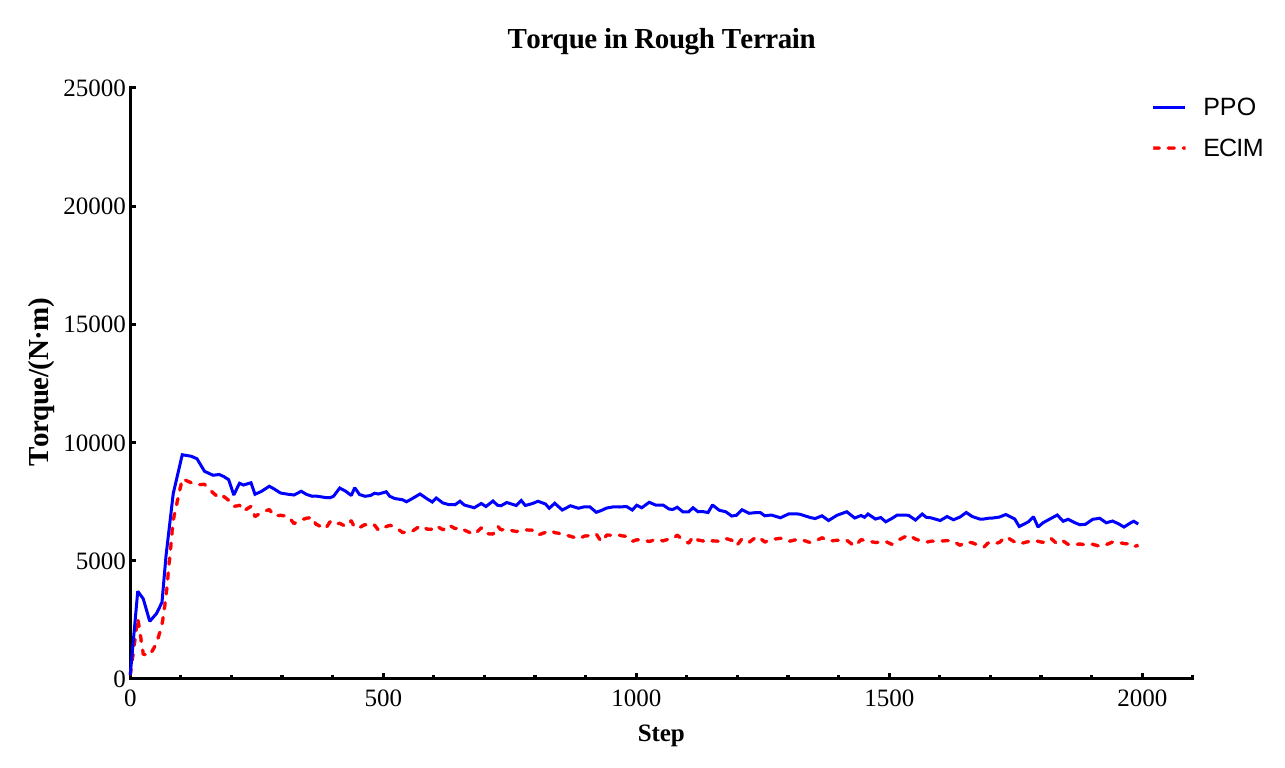}}}\hspace{5pt}
  
\subfloat[Torque in Stair-Climbing Terrain]{%
  \resizebox*{4cm}{!}{\includegraphics{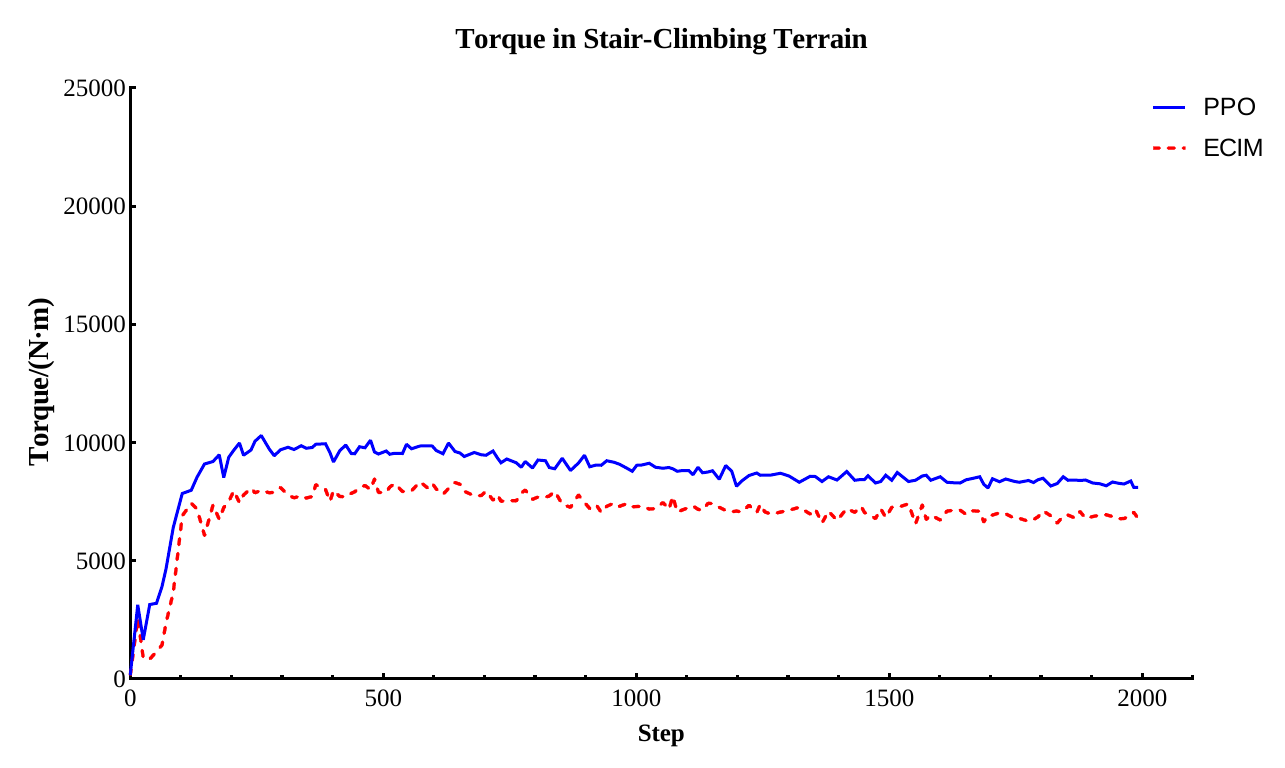}}}\hspace{5pt}
\subfloat[Torque in Stair-Descending Terrain]{%
  \resizebox*{4cm}{!}{\includegraphics{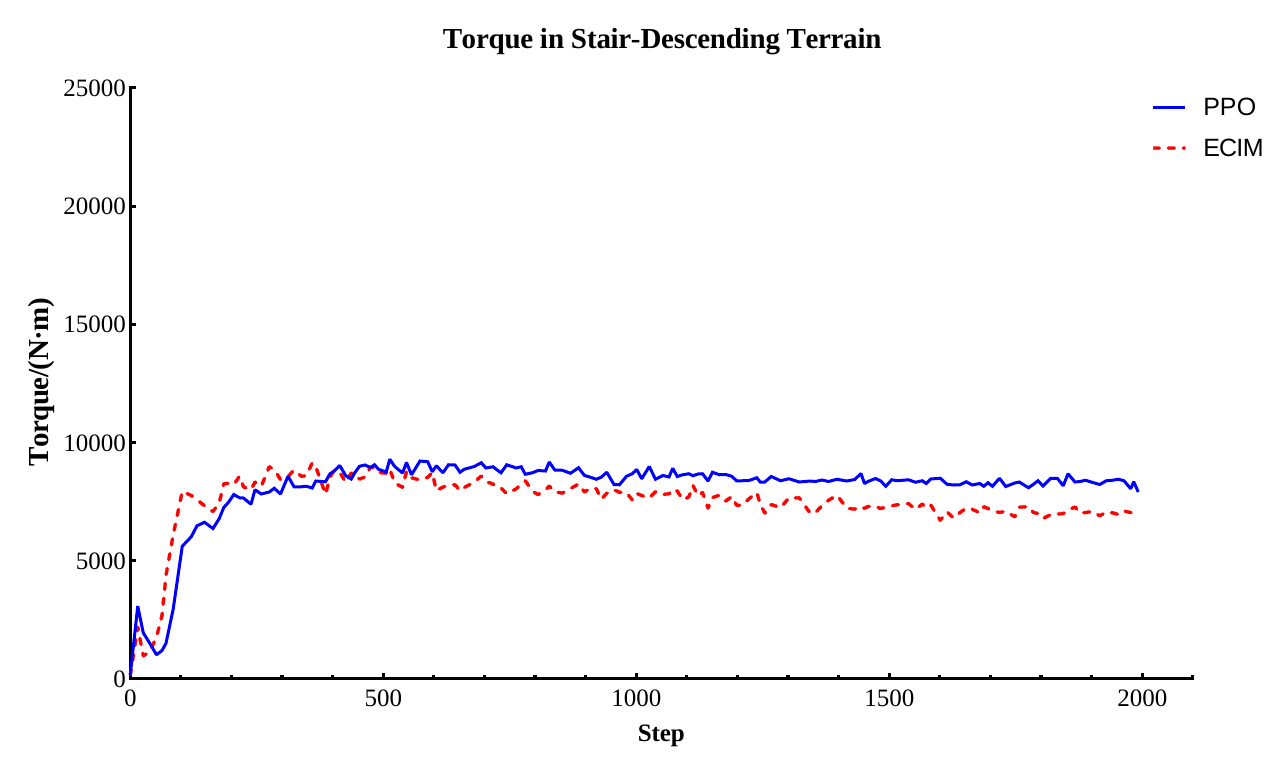}}}\hspace{5pt}
\subfloat[Torque in Stepping Stones Terrain]{%
  \resizebox*{4cm}{!}{\includegraphics{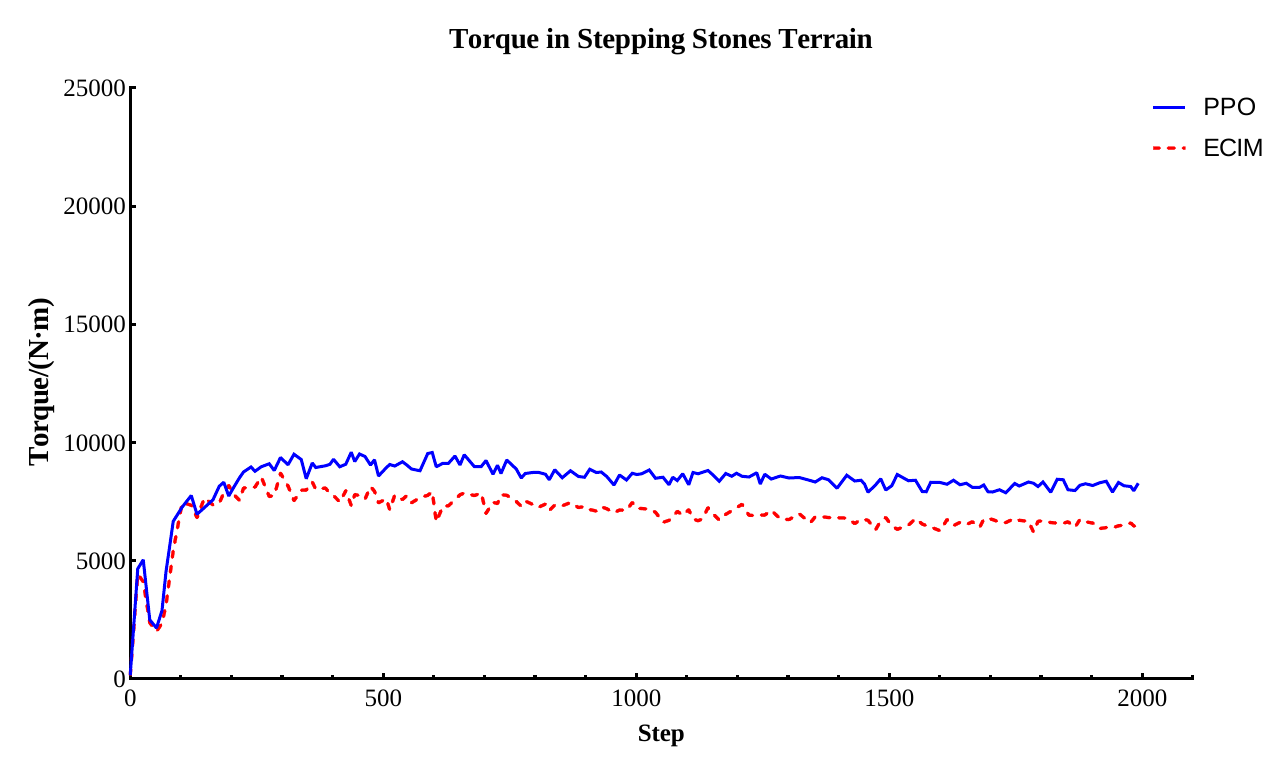}}}\hspace{5pt}
\caption{Torque variations under six different terrain environments for PPO and ECIM.}
\label{fig:torque}
\end{figure}

Figure~\ref{fig:torque} shows the joint torque evolution of PPO and ECIM across the six terrain environments. Joint torque reflects the actuation effort of the robot and is closely related to energy usage and hardware load. As summarized in Table~\ref{tab:performance_summary}, ECIM achieves a reduction of approximately $11.3\%$--$20.0\%$ in Torque RMS across all terrains, with an overall reduction of about $15.8\%$. This means that, for similar or better task performance, ECIM needs less torque on average than PPO.
In the early stage of training, both methods quickly reduce torque from high initial values as basic locomotion skills are learned. After this phase, the PPO torque curves tend to flatten and still show noticeable fluctuations, indicating that the policy remains more aggressive and less consistent in its torque usage. In contrast, ECIM maintains a steady downward trend and eventually converges to lower and more stable torque levels on all terrains. This behaviour is consistent with the design of IMDEEM, which provides an intrinsic reward signal that encourages efficient exploration and discourages unnecessarily strong control actions.
On the complex terrains, such as stairs and stepping stones, this difference becomes especially important. ECIM is able to keep torque levels low at the same time still achieving robust and stable locomotion, but PPO often relies on higher and more variable torque to cope with the same challenges. Combined with the improvements in reward, pitch stability and joint acceleration discussed above, these results show that the proposed ECIM framework achieves more energy-efficient and hardware-friendly control than PPO, which is crucial for long-term deployment of quadruped robots in real-world environments.

\subsection{Dissecting ECIM: Ablations of Entropy Control, Smoothness, and Intrinsic Motivation}

To better understand the contribution of each component in the ECIM framework, we run a set of ablation experiments on the simulated quadrupedal robot across the six terrains. We compare the full ECIM model with three variants where one module is removed---ECIM\_minus\_AECPOM, ECIM\_minus\_MCRF, and ECIM\_minus\_IMDEEM---and with the PPO baseline. The evaluation focuses on \texttt{Steps-to-R*}, \texttt{Pitch RMS}, \texttt{Acc RMS}, and \texttt{Torque RMS}. Our policy will be better when this are lower. Together, these metrics capture training efficiency, motion stability and smoothness, and energy consumption. Detailed results for all terrains are reported in Tables~\ref{tab:ppo_results}--\ref{tab:ECIM_minus_imdeem}.

\subsubsection{Baseline Comparison: PPO vs Full Entropy-Controlled Intrinsic Motivation}

\begin{table}[htbp]
\centering
\caption{Experimental results of PPO.}
\label{tab:ppo_results}
\begin{tabular}{lcccccc}
\hline
Metric & Flat & Slope & Rough & Stairs Up & Stairs Down & Stepping Stones \\
\hline
Steps-to-R*  & 1241.42 & 1207.90 & 1389.98 & 1397.47 & 1207.59 & 1372.31 \\
Pitch RMS    & 0.0862  & 0.0905  & 0.0840  & 0.0967  & 0.0847  & 0.0877  \\
Acc RMS      & 0.0697  & 0.0871  & 0.0874  & 0.0762  & 0.0760  & 0.0649  \\
Torque RMS   & 0.1020  & 0.0930  & 0.0827  & 0.1151  & 0.1010  & 0.1116  \\
\hline
\end{tabular}
\end{table}

\begin{table}[htbp]
\centering
\caption{Experimental results of ECIM (full model).}
\label{tab:ECIM}
\begin{tabular}{lcccccc}
\hline
Metric & Flat & Slope & Rough & Stairs Up & Stairs Down & Stepping Stones \\
\hline
Steps-to-R*  & 876.45  & 839.21  & 836.39  & 838.03  & 871.97  & 859.18  \\
Pitch RMS    & 0.0536  & 0.0520  & 0.0546  & 0.0540  & 0.0511  & 0.0584  \\
Acc RMS      & 0.0593  & 0.0572  & 0.0571  & 0.0562  & 0.0574  & 0.0532  \\
Torque RMS   & 0.0612  & 0.0669  & 0.0720  & 0.0679  & 0.0788  & 0.0608  \\
\hline
\end{tabular}
\end{table}

As shown in Table~\ref{tab:ppo_results}, the vanilla PPO baseline performs poorly across all environments. It is worse than other algorithm. The average \texttt{Steps-to-R*} exceeds 1300 steps, and both \texttt{Pitch RMS} and \texttt{Torque RMS} are high, indicating inefficient exploration, unstable locomotion, and higher actuation effort. In contrast, the full ECIM model in Table~\ref{tab:ECIM} achieves clear improvements:
\begin{itemize}
    \item the average \texttt{Steps-to-R*} is reduced to $853.5$, a $35.2\%$ decrease compared to PPO;
    \item \texttt{Pitch RMS} drops from $0.0883$ (PPO average) to $0.0548$, a $37.9\%$ reduction, indicating much better posture stability;
    \item \texttt{Torque RMS} decreases from $0.1011$ (PPO average) to $0.0679$, a $32.8\%$ reduction, reflecting more energy-efficient actuation.
\end{itemize}
These results confirm that the full ECIM framework, which combines AECPOM, MCRF and IMDEEM, enables faster learning, more stable motion and lower energy usage across all terrains.

\subsubsection{Removing Adaptive Entropy Control (AECPOM): Effects on Exploration and Generalization}

\begin{table}[htbp]
\centering
\caption{Experimental results of ECIM\_minus\_AECPOM.}
\label{tab:ECIM_minus_aecpom}
\begin{tabular}{lcccccc}
\hline
Metric & Flat & Slope & Rough & Stairs Up & Stairs Down & Stepping Stones \\
\hline
Steps-to-R*  & 1095.56 & 1049.01 & 1045.49 & 1047.53 & 1089.96 & 1073.98 \\
Pitch RMS    & 0.0552  & 0.0536  & 0.0562  & 0.0557  & 0.0526  & 0.0602  \\
Acc RMS      & 0.0610  & 0.0589  & 0.0588  & 0.0579  & 0.0591  & 0.0548  \\
Torque RMS   & 0.0630  & 0.0689  & 0.0742  & 0.0699  & 0.0811  & 0.0626  \\
\hline
\end{tabular}
\end{table}

Removing the adaptive entropy-controlled policy optimization mechanism (AECPOM) leads to a clear drop in learning efficiency, as shown in Table~\ref{tab:ECIM_minus_aecpom}. The average \texttt{Steps-to-R*} increases to $1068.5$, which is about $+215.0$ steps compared to the full ECIM model, and the degradation is most pronounced on complex terrains such as rough and stepping-stones. In contrast, stability- and energy-related metrics (\texttt{Pitch RMS}, \texttt{Acc RMS}, and \texttt{Torque RMS}) remain relatively close to those of the full model. This indicates that AECPOM mainly acts on the exploration-exploitation balance during training, helping the policy escape local optima and reach high reward levels more quickly, especially in difficult environments. Without AECPOM, the agent can still obtain reasonably stable and efficient gaits, but it takes longer to learn them and generalizes less robustly across terrains.

\subsubsection{Removing Motion Continuity (MCRF): Effects on Locomotion Stability and Smoothness}

\begin{table}[htbp]
\centering
\caption{Experimental results of ECIM\_minus\_MCRF.}
\label{tab:ECIM_minus_mcrf}
\begin{tabular}{lcccccc}
\hline
Metric & Flat & Slope & Rough & Stairs Up & Stairs Down & Stepping Stones \\
\hline
Steps-to-R*  & 942.38  & 915.67  & 912.45  & 918.21  & 950.73  & 936.50  \\
Pitch RMS    & 0.0618  & 0.0589  & 0.0621  & 0.0615  & 0.0592  & 0.0665  \\
Acc RMS      & 0.0635  & 0.0612  & 0.0610  & 0.0598  & 0.0619  & 0.0573  \\
Torque RMS   & 0.0648  & 0.0701  & 0.0758  & 0.0715  & 0.0829  & 0.0642  \\
\hline
\end{tabular}
\end{table}

The ECIM\_minus\_MCRF variant in Table~\ref{tab:ECIM_minus_mcrf} shows a different pattern. The average \texttt{Steps-to-R*} remains relatively low at $931.0$, indicating that task completion speed is still good. However, \texttt{Pitch RMS} increases from $0.0548$ (full ECIM) to $0.0620$, and \texttt{Acc RMS} increases from $0.0567$ to $0.0608$, which points to degraded motion smoothness and dynamic stability. In practice, the robot exhibits more oscillatory and jerky movements, particularly on sloped and stair-descending terrains, even though it still reaches the target reward threshold in a reasonable number of steps. This confirms that the motion continuity regularization framework (MCRF) is the main driver for stabilizing motion dynamics and enforcing consistent locomotion patterns across terrain transitions, without sacrificing convergence speed.

\subsubsection{Removing Intrinsic Motivation–Driven Exploration (IMDEEM): Effects on Energy Use}

\begin{table}[htbp]
\centering
\caption{Experimental results of ECIM\_minus\_IMDEEM.}
\label{tab:ECIM_minus_imdeem}
\begin{tabular}{lcccccc}
\hline
Metric & Flat & Slope & Rough & Stairs Up & Stairs Down & Stepping Stones \\
\hline
Steps-to-R*  & 903.15  & 878.44  & 875.22  & 880.99  & 908.66  & 894.33  \\
Pitch RMS    & 0.0563  & 0.0541  & 0.0568  & 0.0562  & 0.0534  & 0.0610  \\
Acc RMS      & 0.0602  & 0.0581  & 0.0579  & 0.0570  & 0.0583  & 0.0540  \\
Torque RMS   & 0.0785  & 0.0832  & 0.0891  & 0.0854  & 0.0963  & 0.0772  \\
\hline
\end{tabular}
\end{table}

When the intrinsic motivation-driven exploration enhancement mechanism (IMDEEM) is removed, Table~\ref{tab:ECIM_minus_imdeem} shows that \texttt{Torque RMS} increases sharply to an average of $0.0849$, which is $+0.0170$ higher than the full ECIM model, corresponding to a $25.0\%$ increase. In contrast, \texttt{Steps-to-R*} ($885.0$ on average) and the stability-related metrics remain close to those of the full model. It shows IMDEEM play an important role in energy efficiency. The agent can still learn effective gaits and converge quickly, but it does so with higher actuator effort when IMDEEM is absent. For real-world deployment scenarios with limited power budgets and hardware wear concerns, such additional energy consumption is undesirable, highlighting the importance of IMDEEM for practical quadruped locomotion.

\subsubsection{Model Attribution and Cross-Terrain Generalization Robustness}

To quantify the relative contributions of AECPOM, MCRF and IMDEEM on the key performance metrics, we compute the Attribution Gain (AG) for each module. Also, the added AG values across five core metrics are summarized in Table~\ref{tab:attribution_gain}. The results show that:
\begin{itemize}
    \item AECPOM achieves the highest AG values in \texttt{Reward} (+1.1008) and \texttt{Steps-to-R*} ($-213.3833$), indicating its dominant role in improving task performance and accelerating policy convergence;
    \item MCRF contributes most strongly to \texttt{Pitch RMS} ($-0.0077$) and \texttt{Acc RMS} ($-0.0040$), highlighting its impact on motion smoothness and dynamic stability;
    \item IMDEEM shows the largest AG in \texttt{Torque RMS} ($-0.0170$), far exceeding the other modules, which confirms its leading role in energy efficiency.
\end{itemize}
These quantitative results make the specialized effect of each component explicit for desgin adjustments. Also it support the ECIM functional decoupling and role specialization of each module inside.

\begin{table}[htbp]
\centering
\caption{Attribution Gain (AG) of each ECIM module on key metrics.}
\label{tab:attribution_gain}
\begin{tabular}{lccccc}
\hline
Module & Reward & Steps-to-R* & Pitch RMS  & Acc RMS  & Torque RMS \\
\hline
AECPOM   & \textbf{+1.1008}  & \textbf{-213.3833} & -0.0016 & -0.0017 & -0.0020 \\
MCRF     & +0.7048  & -75.7850 & \textbf{-0.0077} & \textbf{-0.0040} & -0.0036 \\
IMDEEM   & +0.3832  & -36.5933 & -0.0024 & -0.0008 & \textbf{-0.0170} \\
\hline
\end{tabular}
\end{table}

We further evaluate the consistency of each model's performance across terrains by computing the standard deviation (Std) of each metric over the six environments, as shown in Table~\ref{tab:std_deviation}. ECIM attains the lowest standard deviations on all metrics (e.g., \texttt{Steps-to-R*}: 16.51, \texttt{Reward}: 0.55), indicating better cross-terrain stability and generalization. The ablated variants show slightly higher variance: ECIM\_minus\_AECPOM has larger fluctuations in \texttt{Steps-to-R*}, ECIM\_minus\_MCRF has marginally higher \texttt{Pitch RMS} Std, and ECIM\_minus\_IMDEEM exhibits increased \texttt{Torque RMS} Std. Although the absolute differences are rather small, the full ECIM model is consistently more stable than all ablated variants across all metrics.

\begin{table}[htbp]
\centering
\caption{Standard deviation of performance across environments (lower is better).}
\label{tab:std_deviation}
\begin{tabular}{lccccc}
\hline
Model & Steps-to-R* & Pitch RMS & Acc RMS & Torque RMS & Reward \\
\hline
PPO                  & 84.8851 & 0.0043 & 0.0083 & 0.0109 & 2.1232 \\
ECIM                 & 16.5145 & 0.0023 & 0.0018 & 0.0062 & 0.5518 \\
ECIM\_minus\_AECPOM  & 20.6435 & 0.0024 & 0.0019 & 0.0064 & 0.5242 \\
ECIM\_minus\_MCRF    & 14.5766 & 0.0025 & 0.0019 & 0.0064 & 0.5247 \\
ECIM\_minus\_IMDEEM  & 12.7346 & 0.0024 & 0.0019 & 0.0065 & 0.4922 \\
\hline
\end{tabular}
\end{table}

Overall, the ablation studies show that AECPOM, MCRF and IMDEEM play complementary roles: AECPOM mainly enhances learning efficiency and task performance, MCRF improves motion stability and smoothness, and IMDEEM reduces energy consumption. At the same time, the three modules interact positively and form a coherent pipeline that supports high-performance and robust quadruped locomotion across a wide range of terrains.

\section{Conclusion}

This work presented ECIM, a reinforcement learning framework that combines adaptive entropy exploration, motion continuity regularization and intrinsic reward guidance. The goal is to balance exploitation and exploration at the same time shaping motion quality. In particular, ECIM targets two key limitations of standard deep RL in complex tasks: a trend to get trapped in local optima and a lack of sufficiently deep, structured exploration on more challenging terrain.

Across a set of six terrains with varying difficulty, and using PPO as the baseline, our experiments show that ECIM consistently achieves higher returns and faster convergence while producing smoother and more stable locomotion. The learned gaits exhibit reduced body oscillation, lower joint accelerations and lower torque usage. Together with the ablation study, these results indicate that AECPOM mainly improves learning efficiency and robustness, MCRF is central for stabilizing and smoothing motion, and IMDEEM provides a strong gain in energy efficiency. Importantly, the full ECIM framework improves all three aspects at the same time instead of trading one for another.

Because ECIM encourages richer exploration and includes additional regularization and intrinsic reward terms, each training iteration incurs a slightly higher computational cost than vanilla PPO. However, by using large-scale parallel training in Isaac Gym, the overall wall-clock training time remains practical. For this reason, we focus our evaluation on sample efficiency and behaviour quality rather than absolute runtime.

Although this work focuses on quadruped locomotion with ANYmal, the proposed framework is robust for any exploration core algorthms. The same principles can be applied to other high-dimensional continuous control problems where local optima, unstable motion and energy use are critical concerns. In future work, we plan to extend ECIM to bipedal gait generation, contact-rich robotic arm manipulation and whole-body mobile platforms, and to study sim-to-real transfer under realistic power and hardware constraints. We believe that ECIM offers a reusable way to shape exploration and regularize motion in reinforcement learning, with potential impact well beyond the specific setting studied here.

\begin{con}
\ctitle{Author Contributions}
Wanru Gong and Xinyi Zheng conceived and designed the study. Wanru Gong and Xinyi Zheng conducted data gathering. GH performed statistical analyses. Wanru Gong, Xinyi Zheng,Yuan Hui, Xiaopeng Yang and Xiaoqing Zhu wrote the article.

\ctitle{Financial Support}
This work was supported in part by the National Natural Science Foundation of China under Grant 62103009 and the Natural Science Foundation of Beijing under Grant 4202005.

\ctitle{Conflicts of Interest}
The authors declare that they have no known competing financial interests or personal relationships that could have appeared to influence the work reported in this paper.

\ctitle{Ethical Approval}
Not applicable.

\end{con}
\vspace{5mm} 

\bibliographystyle{ieeetr}
\bibliography{refs}

\end{document}